# Using Self-Supervised Auxiliary Tasks to Improve Fine-Grained Facial Representation

Mahdi Pourmirzaei    Gholam Ali Montazer    Farzaneh Esmaili

{m.poormirzaie, montazer, f.esmaili}@modares.ac.ir

Tarbiat Modares University

## Abstract

Facial emotion recognition (FER) is a fine-grained problem where the value of transfer learning is often assumed. We first quantify this assumption and show that, on AffectNet, training from random initialization with sufficiently strong augmentation consistently matches or surpasses fine-tuning from ImageNet. Motivated by this result, we propose Hybrid Multi-Task Learning (HMTL) for FER in the wild. HMTL augments supervised learning (SL) with self-supervised learning (SSL) objectives during training, while keeping the inference-time model unchanged. We instantiate HMTL with two tailored pretext tasks, puzzling and inpainting with a perceptual loss, that encourage part-aware and expression-relevant features. On AffectNet, both HMTL variants achieve state-of-the-art accuracy in the eight-emotion setting without any additional pretraining data, and they provide larger gains under low-data regimes. Compared with conventional SSL pretraining, HMTL yields stronger downstream performance. Beyond FER, the same strategy improves fine-grained facial analysis tasks, including head pose estimation and gender recognition. These results suggest that aligned SSL auxiliaries are an effective and simple way to strengthen supervised fine-grained facial representation without adding extra computation cost during inference time.

## 1. Introduction

Facial expressions are a primary channel for conveying affect and intent in human communication. Roughly one third of communication is verbal, while the remaining two thirds is nonverbal, and facial expressions play a central role among these nonverbal cues [1].

Recent advances in computer vision and deep learning have led to a variety of systems for facial emotion recognition (FER) [1]. Many visual signals can inform emotion analysis, including appearance, gesture, behavior, and scene context. Even so, facial expressions remain the most informative cue for recognizing basic human emotions [2]. Two modeling paradigms are commonly used in FER: the categorical model and the circumplex, or dimensional, model [3].

In the categorical model, Ekman defined a set of basic emotion categories, typically Anger, Happiness, Sadness, Surprise, Disgust, Fear, Contempt, and a Neutral state [4]. The Facial Action Coding System (FACS) represents expressions as combinations of action units that correspond to specific facial muscle movements. FACS provides an objective description of facial activity and is widely used to analyze expressions and infer emotion [4].

The circumplex model of emotion, developed by James Russell [3], represents affect in a two-dimensional circular space with valence and arousal as the axes. The center corresponds to a neutral state on both dimensions, and emotions are positioned around the circle according to their valence and arousal levels [5]. Fig. 1 illustrates this model.

According to the Russell and Ekman models, data labeling for FER requires expert annotators and is not as straightforward as in other computer vision tasks such as object detection. The inherent uncertainty in facial emotion recognition makes annotation difficult and costly, and this uncertainty is even greater for Russell's dimensional model. Consequently, collecting large, high-quality labeled datasets for FER is challenging and can undermine annotation reliability. This limitation reduces the effectiveness of deep learning methods that rely on supervised learning (SL).

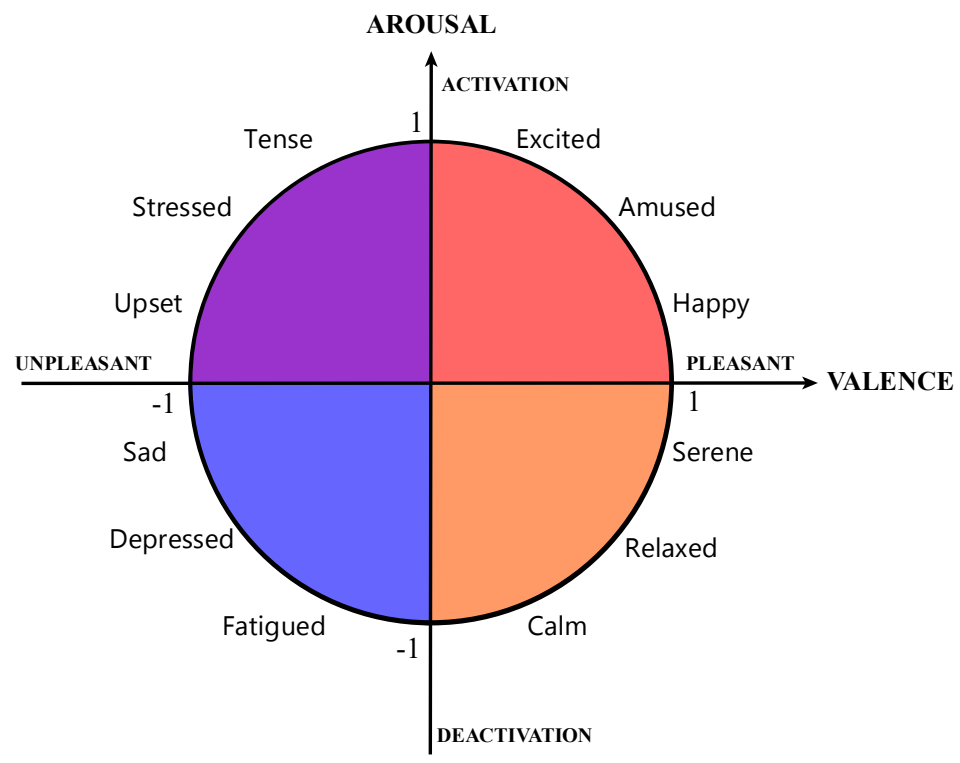

Fig. 1 Circumplex model with arousal and valence axes [3].

Recently, progress in self-supervised learning (SSL) has shown that end-to-end methods can reduce dependence on extensive labels [6–8]. Although contrastive SSL methods are effective, they still have unresolved drawbacks, including the need for large batch sizes [6], substantial computation and data to perform well [6,7,9], and a clear gap behind SL on fine-grained visual tasks [10]. The latter two issues are the main obstacles to improving FER with SSL. To our knowledge, there is no work that seriously investigates SSL for fine-grained FER. Moreover, FER is one of the most challenging fine-grained problems because of label uncertainty. For example, in AffectNet, even with trained human annotators, agreement across emotions was about 60% to 70% [11], which highlights the ambiguity of emotion conveyed through the visual channel.

In some studies [12,13], another family of SSL methods known as pretext task SSL has been used to extract valuable features for downstream tasks, in some cases with pretraining on a single image [14]. In this setting, intermediate layers often produce more useful features than final layers, and linear evaluation across layers tends to rise from low to mid-level layers and then decline toward the top [14].

Since the 1990s, studies have demonstrated the effectiveness of multi-task learning (MTL) in supervised settings [15,16]. Selecting compatible tasks remains nontrivial, because tasks initially benefit from shared representations and may later interfere, a phenomenon often described as cooperation and competition [15]. Viewed through this lens, self-supervised learning (SSL) objectives can act as auxiliary tasks within an MTL framework. This raises our central question: can combining supervised learning (SL) with auxiliary SSL objectives, trained concurrently in MTL, yield better fine-grained FER than SL alone?

In this study, we integrate SSL with SL in an MTL framework, which we call Hybrid Multi-Task Learning (HMTL). HMTL operates only during training: SSL heads are attached to the backbone and removed at test time. The auxiliary SSL objectives encourage the backbone to learn more discriminative features for fine-grained facial representation. Importantly, HMTL is not a pretraining strategy. In other words, SSL can be used in two ways, either to pretrain model weights or as an auxiliary objective alongside SL. We adopt the latter and show its advantages. Throughout this work, we use HMTL, SL+SSL, and SL with an auxiliary SSL task terms interchangeably.

The contributions of this work are summarized as follows:

1. We quantify the effect of augmentation strength in FER when training from scratch versus fine-tuning ImageNet-pretrained weights. With sufficiently strong augmentation, training from random initialization outperforms fine-tuning under any augmentation regime.
2. We perform SSL pretraining with several pretext tasks and show that two proposed SSL variants learn more useful features for FER. Both are also effective as standalone pretraining steps.
3. We formulate and test the hypothesis that adding appropriate SSL auxiliary tasks to SL within an MTL framework improves the downstream supervised objective.
4. On AffectNet, HMTL delivers substantial gains for both dimensional and categorical emotion recognition across all augmentation levels, including settings that use only 20% of the highly imbalanced training set.
5. We evaluate auxiliary SSL objectives on two additional fine-grained facial tasks, head pose estimation and gender recognition, and observe a substantial reduction in average head pose error.

# 2. Related works

Methods for facial emotion recognition (FER) are commonly grouped into conventional pipelines and end-to-end learning approaches, with the latter achieving stronger results in many vision tasks [1].

Deep learning now underpins most FER systems, and convolutional neural networks (CNNs) have been central for years [1]. CNNs reduce the need for manual preprocessing by supporting end-to-end learning from raw inputs [17]. Some studies, however, do not rely solely on CNN features. For example, [18], combines features learned by CNNs through pretraining with handcrafted descriptors summarized using a bag-of-visual-words representation.

End-to-end methods have generally surpassed traditional pipelines and have focused on improved network design [19–21]. In [19], manifold networks combined with CNNs and covariance pooling outperform CNNs paired with a Softmax classifier. Another study [20] proposed two CNN-based models that vary kernel sizes and the number of filters for FER.

Video-based FER extends these ideas to temporal data [22]. A common setup classifies frames into emotion categories such as happy or sad. Although Inception and ResNet architectures have shown strong results, they do not natively aggregate temporal information. To address this gap, a three-dimensional Inception-ResNet architecture was introduced to learn spatiotemporal features across frame sequences [23]. In this type of architecture, geometric and temporal features were extracted in a sequence of frames with the three-dimensional model. Another study [24], employs two complementary CNNs for videos, one capturing temporal appearance characteristics and the other modeling temporal geometric dynamics of facial landmark points.

Beyond pure CNNs, combining CNNs with recurrent models such as GRUs and LSTMs can further improve video-based FER. In [25], three architectures integrate CNN features with RNNs to recognize dimensional emotions in an MTL setting. Frames are first processed by a CNN, then multi-level features are fed to multiple RNNs. Integrating LSTMs with CNNs enables variable-length inputs and outputs, which is a key advantage for video analysis [1]. A hybrid LSTM-CNN model in [26] was shown to outperform prior 3D-CNN approaches over time.

Attention mechanisms have also delivered notable gains [27–30]. A CNN with visual attention can emphasize informative regions for feature extraction and expression detection [27]. An attention-augmented CNN in [31], handles facial occlusions and introduces Patch-Gated CNNs that automatically focus on unoccluded regions for end-to-end expression recognition.

Finally, several works explore alternative architectures and formulations. Graph convolutional networks have been applied by constructing undirected facial graphs [32]. An identity-free conditional GAN (IF-GAN) in [33] seeks to disentangle identity from expression, suppressing identity-related cues so that expression-specific features can drive emotion classification.

## 2.2. Self-supervised learning

SSL methods learn representations from unlabeled data without human-provided annotations. As a subset of unsupervised learning, SSL reduces the cost of collecting and labeling large datasets. In SL, models learn from paired data (x, y) where y is annotated by humans. In SSL, supervision is derived automatically from $x$, so labels $y$ are generated from the inputs themselves [34]. SSL has proved useful at multiple scales, including robustness to adversarial perturbations, label corruption, and semi-supervised learning.

Many SSL approaches have been proposed for computer vision [34] which can be grouped into three categories:

1. Contrastive learning
2. Non-contrastive learning
3. Pre-text task learning

In the mid-2010s, pretext task learning became popular, including colorization, inpainting, and jigsaw puzzles [35,36]. More recently, contrastive and non-contrastive SSL methods have achieved strong results on challenging benchmarks such as ImageNet [6,8,9]. Despite these advances, such methods often require substantial data and compute to perform well, and their training procedures can be difficult. Moreover, they tend to lag behind SL on fine-grained problems such as FER, which indicates room for improvement [10].

Among pretext techniques, random rotation is widely used. In [12], 2D rotations are applied to input images and a CNN predicts the rotation angle, which encourages directional sensitivity. Jigsaw-based SSL has also been explored: [37,38] partition each image into tiles, shuffle them, and train a Siamese network to recover spatial order. In addition, [13] combines multiple pretext tasks by jointly solving jigsaw puzzles, inpainting, and colorization.

## 2.3. Imbalance dataset

In classification, class imbalance occurs when samples are unevenly distributed across classes. Networks then learn overrepresented classes more effectively, which degrades performance on underrepresented classes. Severe imbalance can bias the classifier toward the dominant class and reduce overall accuracy [39]. Common strategies to address this issue include:

1. Up sampling
2. Down sampling
3. Customize loss function

Customized loss functions have shown strong performance in FER [11]. In this work, we consider weighted loss and focal loss as two popular choices. Weighted loss assigns larger weights to samples from minority classes to counteract imbalance. Focal loss [40] down-weights well-classified examples by modulating the loss with adjustable $\alpha$ and $\gamma$ coefficients. As prediction confidence increases, the contribution to the loss decreases, so the model focuses more on hard and misclassified examples.

# 3. Method

We define three augmentation regimes (Appendix A) to study their impact on FER. First, we compare fine-tuning from ImageNet initialization with training from scratch under each augmentation level. Next, we evaluate features learned by SSL pretraining. Finally, we introduce our Hybrid Multi-Task Learning (HMTL) architecture for FER.

## 3.1. Supervised learning approach

To compare augmentation settings, we train models with either ImageNet-pretrained weights or random initialization (Fig. 2). EfficientNet-B0 and EfficientNet-B2 serve as backbones [41].

This yields six training modes: No augment w/ random initialization - No augment w/ ImageNet initialization - Weak augment w/ random initialization - Weak augment w/ ImageNet initialization - Strong augment w/ random initialization - Strong augment w/ ImageNet initialization

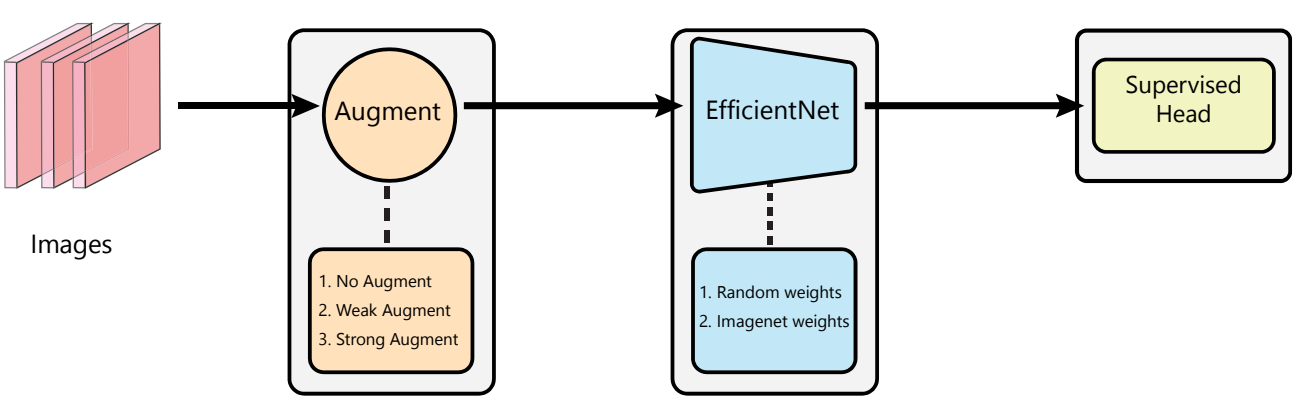

Fig. 2 Supervised learning (SL) pipeline under different augmentation levels and initialization types.

## 3.2. Self-supervised learning approach

We consider three SSL objectives: rotation, puzzling, and inpainting.

**Rotation.** Each image is rotated by $N \times 45°$, where $N \in \{0, \ldots, 7\}$. The task is an eight-way classification problem that predicts the rotation index.

**Puzzling.** Our customized pretext task differs from conventional jigsaw approaches that process tiles with a Siamese network. As illustrated in Fig. 3, an image is partitioned into $N$equal square regions, where $N$is a perfect square (for example, 4 or 9). The tiles are randomly shuffled and reassembled into a single composite image. The network then predicts the original location of each tile using separate classification heads, one per region.

**Inpainting.** We study two variants. (1) A one-stage reconstruction objective that uses a pixel-wise loss (PWL), such as mean squared error (MSE). We employ this variant for both pretraining and HMTL. (2) A perceptual-loss variant used within HMTL. Here, we add a FER perceptual loss (PL) computed in the feature space of a fixed teacher network that was trained in a standard supervised manner on the same dataset. The PL encourages the decoder to reconstruct the missing content so that the resulting representation matches that of the original unmasked image (Fig. 4). In this variant, we use only the PL without a pixel-wise term.

For the in-painting objective, we first localize a face region of interest (ROI) that covers the eyes, nose, and mouth using dataset landmarks when available, otherwise a standard face detector. The ROI is expanded by a small margin to preserve local context (10 to 15 percent of the box size). During training, each image is masked with probability $p_{\text{erase}}$. When masking is applied, we sample a square patch of fixed side length $s$relative to the ROI, for instance $s = r \cdot \min(W_{\text{ROI}}, H_{\text{ROI}})$ with a constant ratio $r$. The patch center is drawn uniformly inside the ROI and resampled if any side would fall outside the ROI so that the entire square remains within facial attributes. The erased region is filled using per-channel dataset means rather than zeros to stabilize optimization under pixel-wise losses. For some experiments we create $K \in \{1,2\}$ independent masks per image and ensure that at least one intersects a high-saliency subregion around the eyes or mouth by using soft landmark heatmaps. The masked image and the corresponding binary mask are fed to the decoder. In the one-stage variant, training minimizes a pixel-wise loss on the masked area and leaves unmasked pixels unchanged. In the perceptual-loss variant, the decoder output is scored only in the feature space of the fixed teacher network using activations from mid-level layers, again restricted to the masked area. To avoid shortcut learning, we apply global geometric and color augmentations before sampling the mask, and we keep the same mask for all augmented views of a given image within a mini-batch. Fig. 4a illustrates the ROI definition and Fig. 4b shows a sampled square within this region.

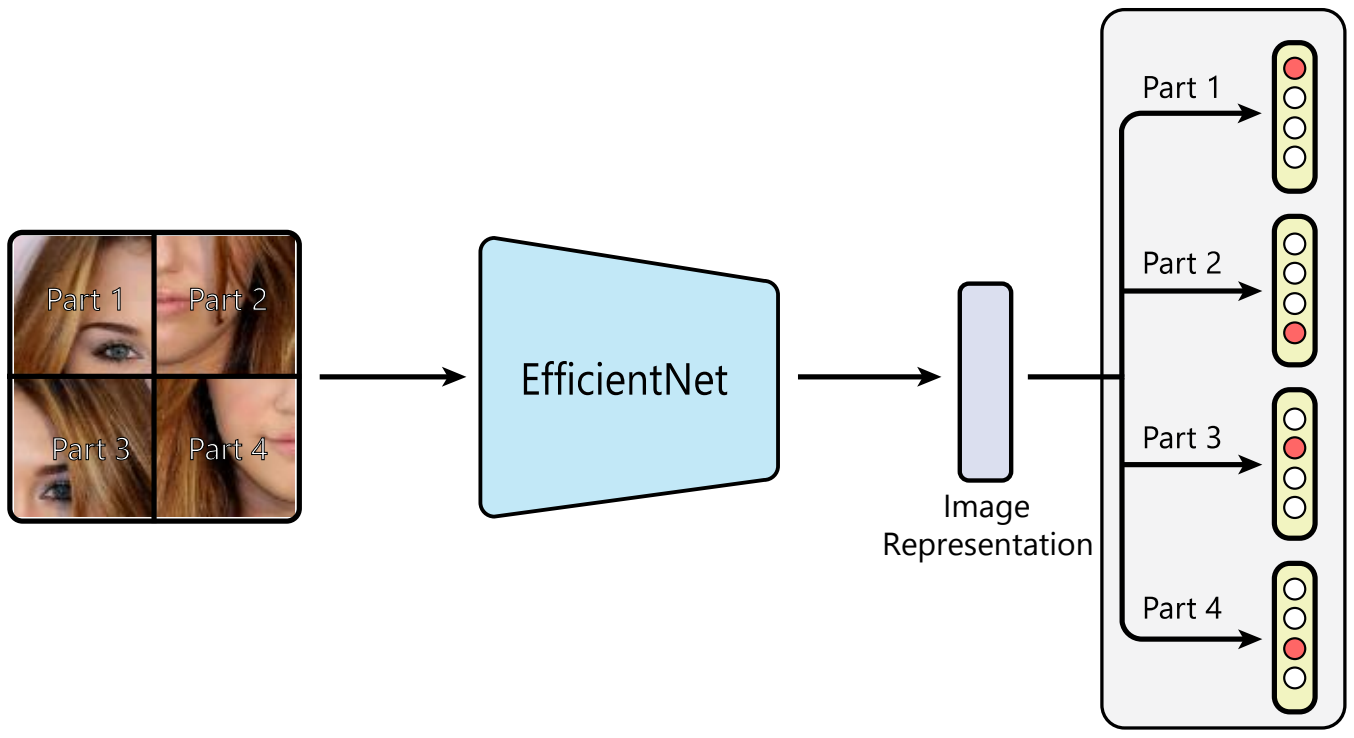

Fig. 3 SSL puzzling procedure. An image is split into tiles, the tiles are shuffled, then reassembled into a single composite. During training, the puzzled image is the input and the label for each tile is its original location. In the example shown, tile one maps to region one, tile two to region four, tile three to region two, and tile four to region three.

## 3.3. Hybrid multi-task learning approach (adding auxiliary SSL)

The impact of multitask learning in neural networks is well established [16]. In computer vision domain, images contain latent cues that can indirectly improve recognition performance for a range of different tasks, yet these cues are not directly optimized by standard SL objectives. For example, jointly recognizing facial structure and expression demonstrated by recent works that can enhance emotion recognition [42][43]. We argue that many of this latent information for fane-grained facial representation is not sufficiently exposed by emotion labels alone and therefore we need additional signals, like labels of other tasks to provide useful cues for a target task like FER.

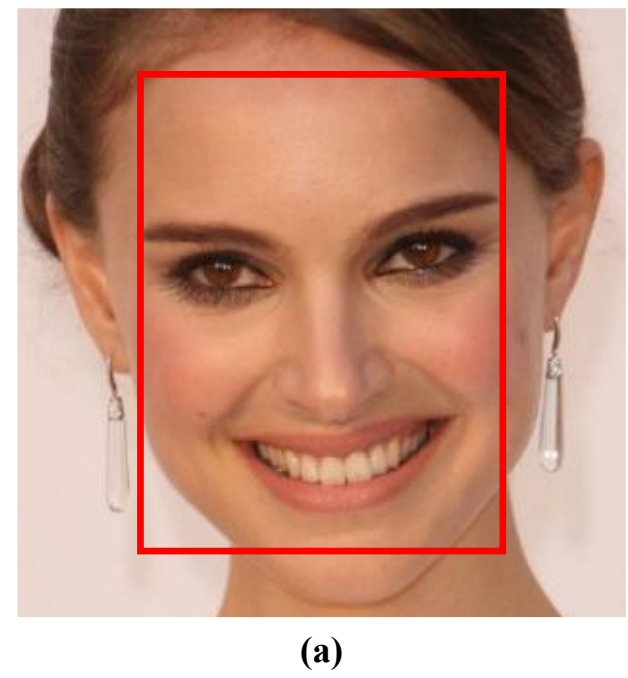

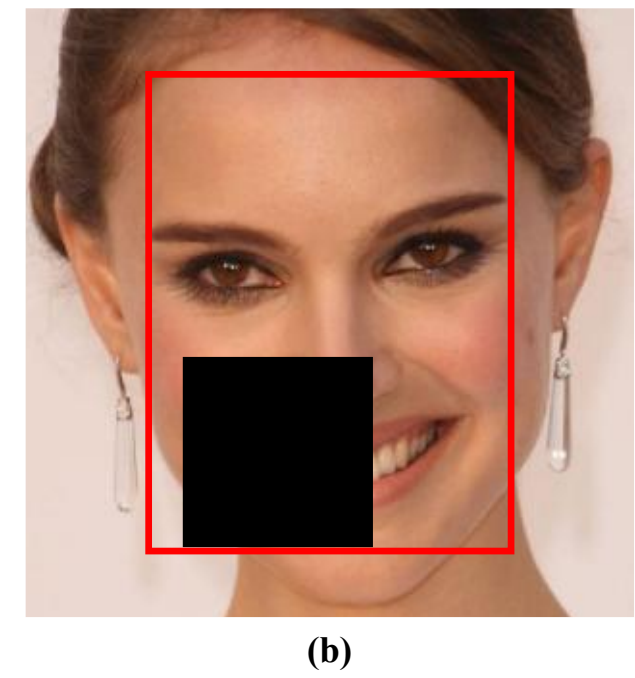

**(a)** **(b)**

Fig. 4 Proposed two-stage inpainting with a perceptual-loss (PL) pretext. (a) The "original image" is used to compute the decoder loss. A fixed rectangular region marks an expression-relevant facial area. (b) A cutout version of the image is created for training. The decoder head reduces the distance between the offline encoder representations of the inpainted image and the original image.

Therefore, given that collecting multi-label annotations in FER (for example, AUs and landmarks) is expensive and difficult, we propose the following hypothesis: leveraging appropriate SSL tasks as auxiliary objectives alongside SL in an MTL setting can improve the learned representation for the supervised FER task. Concretely, we attach self-supervised heads (SSHs) to a shared backbone during training and keep only the supervised head (SH) at test time. To select effective auxiliary tasks, each SSL objective is first evaluated in isolation and retained if it yields features that help solve the main FER problem (see Figs. 5 and 6).

$$L_{total} = L_{SL} + \sum_j L_{SSL_j}$$
$$= -\sum_i w_e y_i \, log(\hat{y}_i) - \sum_j \sum_i y_{i,j} \, log(\hat{y}_{i,j}), \qquad (1)$$

Where:

- $L_{SL}$: weighted categorical cross entropy for supervised head
- $L_{SSL}$: categorical cross entropy for puzzling self-supervised heads

$$L_{total} = L_{SL} + L_{SSL} = -\sum_i w_e y_i \, log(\hat{y}_i) - \sum_j \sum_i y_{i,j} \, log(\hat{y}_{i,j})$$
$$-\sum_i y_i \, log(\hat{y}_i) \, , \qquad (2)$$

Where:

- $L_{SL}$: weighted categorical cross entropy for supervised head
- $L_{SSL}$: categorical cross entropy for puzzling self-supervised heads and rotation self-supervised head

$$L_{total} = L_{SL} + L_{Decoder}$$
$$= -\sum_i w_e y_i \, log(\hat{y}_i) - \lambda \sqrt{\frac{1}{n}\sum_{j=1}^{n}\left(\frac{e^{F(\mathrm{I}_{Rec})_j}}{\sum_{k=1}^{n} e^{F(\mathrm{I}_{Org})_k}} - \frac{e^{F(\mathrm{I}_{Org})_j}}{\sum_{k=1}^{n} e^{F(I_{Org})_k}}\right)^2}\,, \qquad (3)$$

Where:

- $L_{SL}$: weighted categorical cross entropy for supervised head
- $L_{Decoder}$: After each representation are given to a softmax layer, the RMSE loss function of two representations will be calculate
- $\lambda$: weight for the decoder head
- $F(I)$: offline model which gets an image, and then outputs a feature representation
- $\mathrm{I}_{Rec}$: reconstructed image by the decoder head
- $\mathrm{I}_{Org}$: original image without cutout

### 3.3.1. Categorical and dimensional models

In the categorical setting, we use a single supervised head (SH) that predicts the eight Ekman emotions. During training, the backbone is optimized to solve both the primary SL objective and the auxiliary SSL objectives described in Equations (1) to (3). At evaluation time, auxiliary pretext operations such as puzzling or cutout are disabled, and only the SH is active.

To couple categorical and dimensional supervision, we define the following loss:

$$L_{Cat-Reg} = \alpha * L_{Cat} + L_{Reg}$$
$$= -\alpha \sum_i y_{Cat_i} \, log\left(\hat{y}_{Cat_i}\right) + \sqrt{\frac{1}{n}\sum_{j=1}^{n}\left(y_{Reg_j} - E(\hat{y}_{Cat})_j\right)^2}$$
$$+ \sqrt{\frac{1}{n}\sum_{j=1}^{n}\left(y_{Reg_j} - E(\hat{y}_{Cat})_j\right)^2}, \qquad (4)$$

Where:

- $L_{Cat}$: categorical cross entropy for categorical head
- $L_{Reg}$: RMSE loss function for arousal and valence heads
- $\hat{y}$: output of softmax layer
- $\alpha$: weight for the categorical head
- $E$: expectation of softmax layer after categorical head's output which calculates a regression value

Figure 7 illustrates the training procedure for the continuous setting (equation 4). The alpha coefficient is considered as a regulator for the attention of the network to the classification part and its value can be changed. This formulation is related to HopeNet [44], which maps continuous targets to both categorical and continuous spaces to stabilize learning.

# 4. Experiments

We evaluated EfficientNet-B0 and EfficientNet-B2 due to their favorable accuracy, parameter efficiency, and FLOPs across benchmarks [41]. Input resolutions were 224×224 for B0 and 260×260 for B2. All models were trained on a single NVIDIA GTX 1080 Ti using the AdaBelief optimizer [45]. The batch size was 64 for most experiments and 32 for inpainting to accommodate the decoder. All implementations used TensorFlow.

## 4.1 AffectNet dataset

We mainly focused on AffectNet is a large-scale FER dataset with more than one million images collected from three search engines using emotion-related queries in multiple languages. Inside AffectNet, there are about 450,000 images annotated in two formats: categorical and dimensional. The categorical set includes 11 labels, eight of which are the basic emotions defined by Ekman. Note that the training set of this dataset is highly imbalanced (see Fig. 8) but the validation set contains 500 images per category. Also, the official test set is not publicly available and therefore, we only reported the results on the validation set [11].

## 4.2. Supervised learning

For SL, we used EfficientNet-B0 as the backbone and attached a linear classifier to predict the eight emotions. The learning rate was 0.0001 for ImageNet fine-tuning mode and 0.001 for training from random initialization mode.

During the experiments of this part, the number of epochs ranged from 20 to 100 depending on augmentation level and initialization. We did not apply weight decay, and we used a step decay schedule for the learning rate. Training setup in the two modes used the AffectNet training set and evaluation to train and evaluate, respectively. Because the training data were strongly imbalanced, we applied weighted cross-entropy to mitigate that issue. We also used dropout and label smoothing with tuned intensities. Each configuration was trained three times with fixed random seeds, and we reported the best validation accuracy among the three runs.

Table 1 and Fig. 9 summarize the six training modes. Based on the results, increasing augmentation strength produced limited gains for the ImageNet fine-tuning mode. In contrast, stronger augmentation substantially benefited from training on random initialization, and even the strong augmentation regime surpassed fine-tuning from ImageNet initialization.

These results indicated that aggressive augmentation was especially effective when random initialization is in place, which closed or exceeded the performance gap to the ImageNet-based fine-tuning mode, while augmentation provided only modest improvements in the fine-tuning setting.

Table 1 Results for SL with different initializations and augmentation levels. All models use an EfficientNet-B0 backbone and are evaluated on the AffectNet validation set.

| Approach | Augment level | Pre-training weights | Accuracy (%) |
| --- | --- | --- | --- |
| SL | No | - | 57.03 |
| SL | Weak | - | 60.09 |
| SL | Strong | - | 60.34 |
| SL | No | ImageNet | 59.3 |
| SL | Weak | ImageNet | 59.57 |
| SL | Strong | ImageNet | 60.17 |

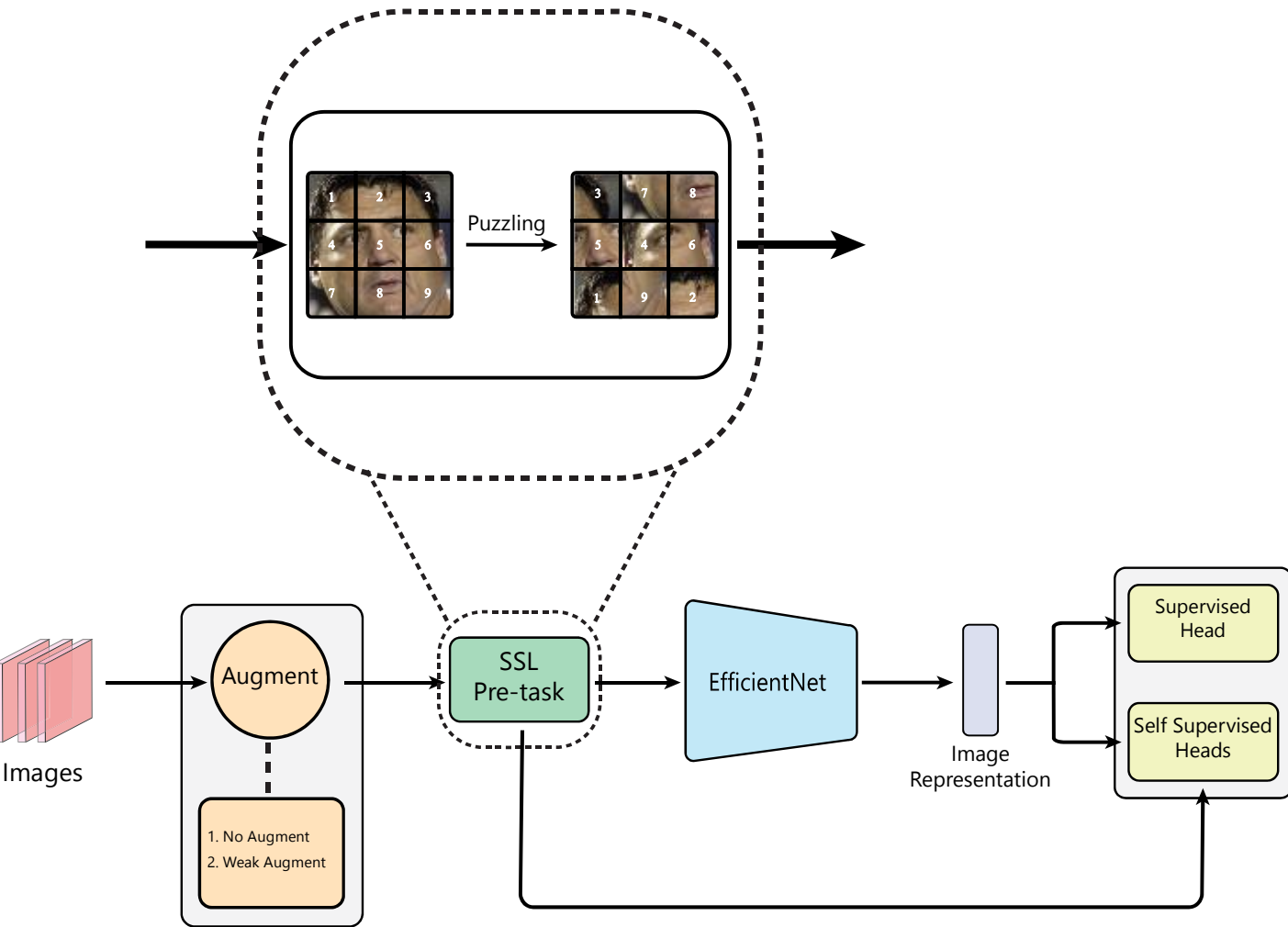

Fig. 5 Training supervised learning with an auxiliary SSL pretext. During training, each input image is partitioned into a 3×3 grid, tiles are shuffled, and the reassembled image is fed to the network. The model learns the main supervised task and a puzzling objective simultaneously. At validation, neither augmentation nor SSL operations are applied. The puzzling block can be replaced with other pretext tasks such as rotation.

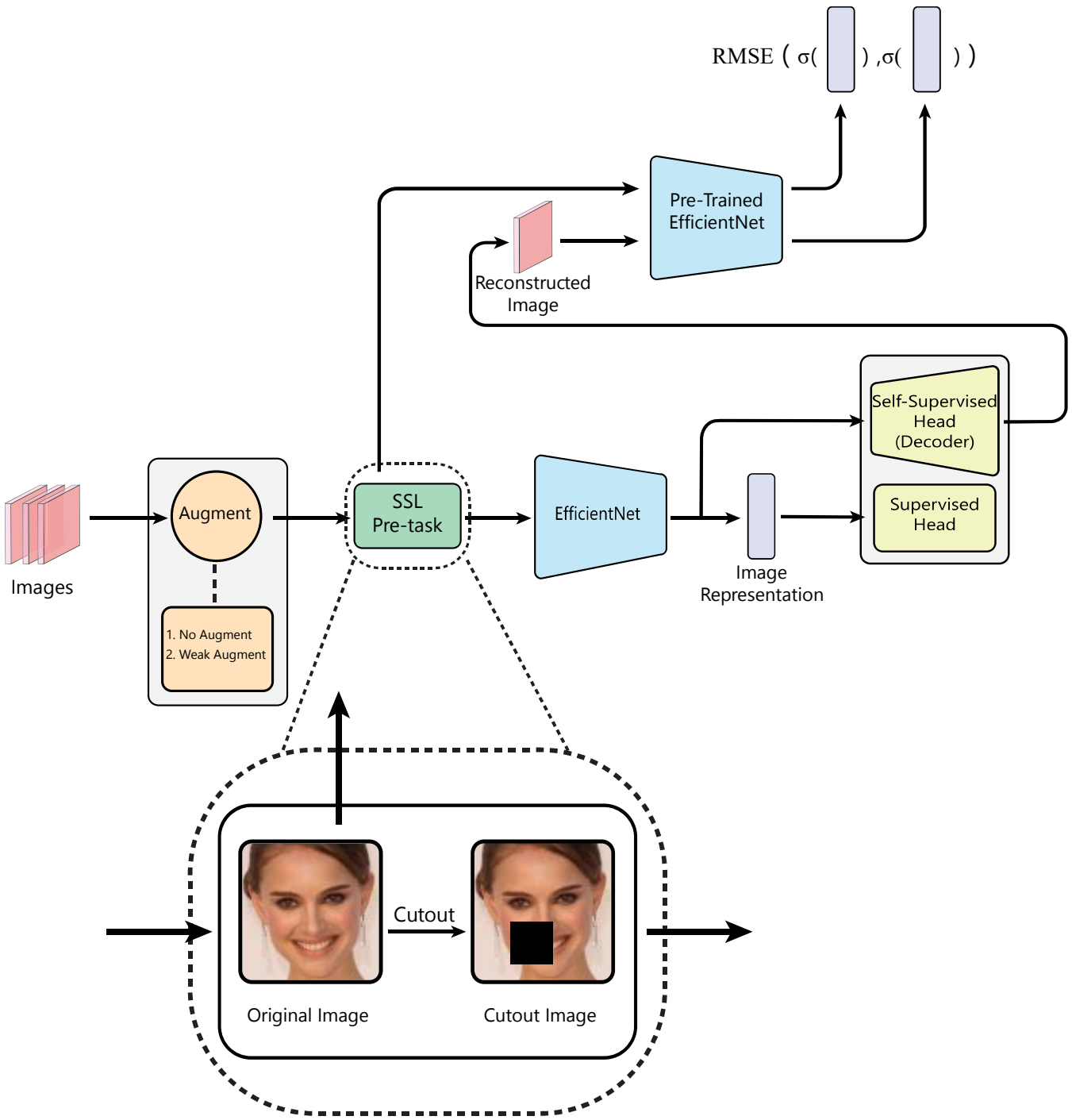

Fig. 6 Two-stage training with an auxiliary decoder and a fixed teacher. Stage one trains an offline backbone from scratch under a chosen augmentation level. Stage two trains a new backbone under the same settings while attaching a decoder head that reconstructs masked inputs. The reconstructed image and the original image are both passed through the offline backbone, and the decoder is optimized so that their feature representations match as closely as possible. The pretrained EfficientNet serves as the teacher and the second EfficientNet serves as the student.

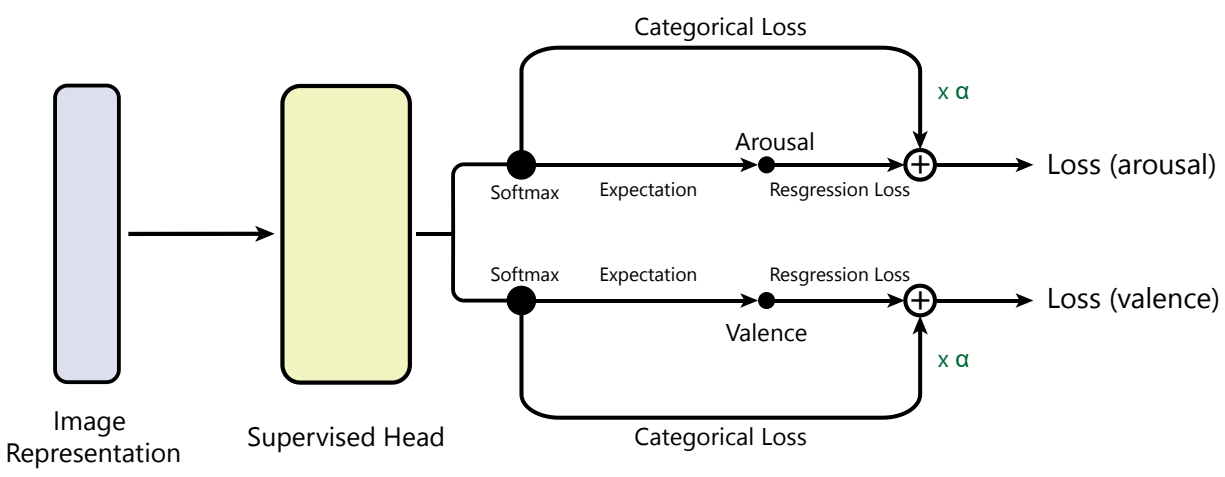

Fig. 7 Coupling categorical and continuous predictions for arousal and valence in the supervised head. The head first predicts a categorical distribution, then converts this distribution to continuous outputs for regression.

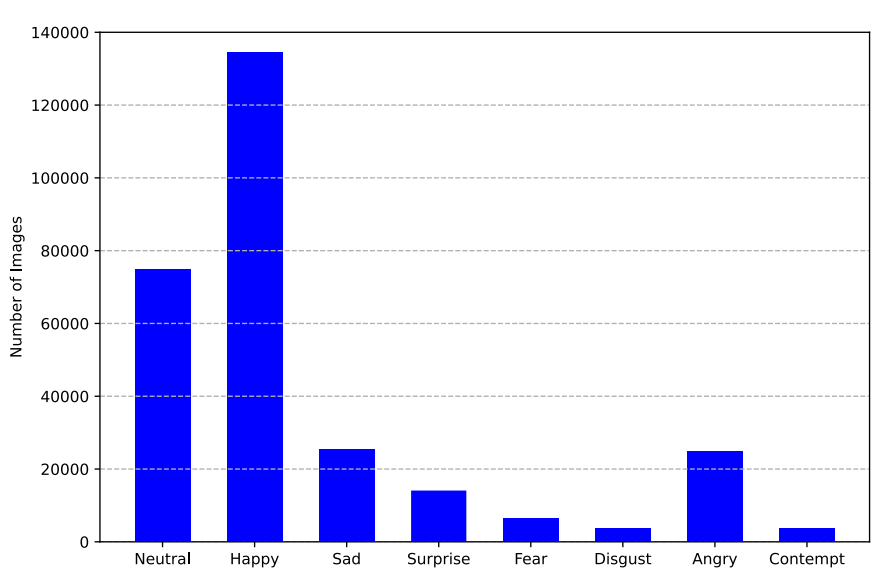

Fig. 8 Distribution of AffectNet labels in the training set. The class frequencies are highly imbalanced.

## 4.3. Self-supervised learning

We evaluated four pretext tasks: (1) puzzling, (2) random rotation, (3) puzzling plus random rotation, and (4) in-painting-pwl. In puzzling, we attached multiple classification heads to the EfficientNet backbone, with the number of heads equal to the number of tiles, for example 4 or 9. In random rotation, we used a single head with eight classes that corresponded to the eight rotation angles. In the combined setting, we joined all puzzling and rotation heads, for example 5 or 10 heads in total. In in-painting-pwl, we attached a convolutional decoder and trained it to reconstruct the original image using a pixel-wise RMSE loss.

Because the first three pretexts were relatively easy for a deep network, we trained them with the strong augmentation regime. For rotation and puzzling-rotation, we removed random rotation from the data pipeline to avoid trivial solutions. For in-painting-pwl, we used no augmentation. As in the supervised experiments, each configuration was trained three times with fixed random seeds.

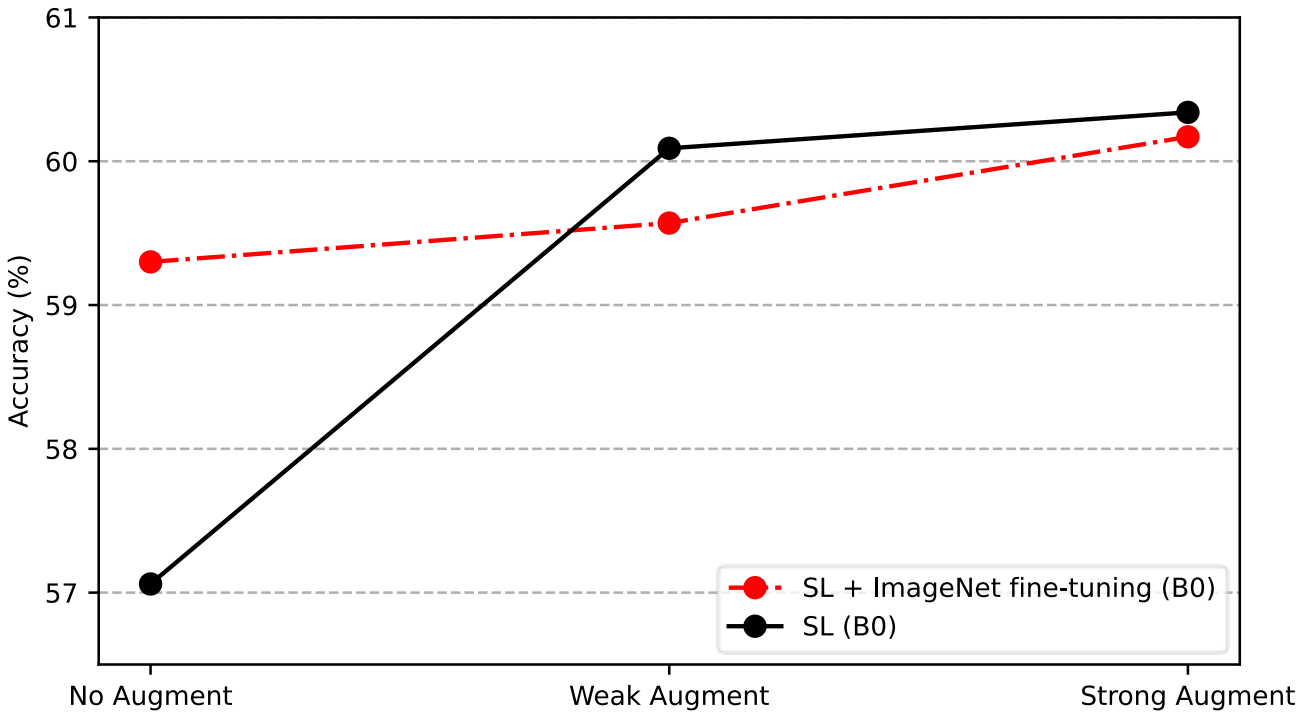

Fig. 9 Comparison of fine-tuning and training from scratch under different augmentation levels.

Although SSL did not require labels, not all AffectNet images contained faces. Among the 450,000 labeled images, about 89,000 were faceless. Among the roughly 550,000 unlabeled images, more than 100,000 were also faceless, which would have introduced noise for FER pretraining. We therefore used the 361,000 labeled images that contained faces for SSL training.

For puzzling losses, we used focal loss to reduce the influence of easy examples and to emphasize hard cases. After training the pretexts to convergence, we removed the SSL heads and evaluated the frozen backbones with a nonlinear classifier on the eight AffectNet emotions. Concretely, we trained a two-layer network on top of each frozen backbone. We also fine-tuned all layers of the SSL-initialized backbones on AffectNet. In both the nonlinear evaluation and fine-tuning settings, we used no augmentation.

Table 2 Nonlinear evaluation on AffectNet with frozen EfficientNet-B0 backbones. Puzzling used a 3×3 grid and rotation used eight angles. In-painting-pwl used a CNN decoder trained with a pixel-wise loss.

| Backbone pre-training weights | Accuracy (%) | Macro F1 |
|---|---|---|
| SSL in-painting-pwl | 34.41 | 0. 3372 |
| SSL puzzling | 32.98 | 0.3227 |
| SSL rotation | 15.48 | 0.1070 |
| SSL puzzling-rotation | 30.08 | 0.2754 |
| Random initialization | 12.5 | 0.04 |

As Table 2 indicated, SSL pretraining produced features that were substantially more informative than random initialization for nonlinear evaluation. In-painting-pwl and puzzling yielded the strongest frozen-feature performance.

Fine-tuning results showed a different pattern (Table 3). Puzzling-rotation improved over random initialization, whereas rotation and in-painting-pwl underperformed in this setting. ImageNet initialization remained a strong baseline for fine-tuning. Together with Table 2, these findings suggested that the utility of SSL pretraining depended on both the chosen pretext and the downstream training protocol.

Table 3 Fine-tuning all layers on the AffectNet training set. The backbone of all approaches is B0. All of the Puzzling methods are 3×3 and all rotation methods are in eight directions.

| Methods | Augment level | Pre-training weights | Accuracy (%) |
|---|---|---|---|
| SL | No | Random initialization | 57.03 |
| SL | No | AffectNet (SSL puzzling) | 57.56 |
| SL | No | AffectNet (SSL rotation) | 54.26 |
| SL | No | AffectNet (SSL puzzling-rotation) | 58.86 |
| SL | No | AffectNet (SSL inpainting-pwl) | 51.84 |
| SL | No | ImageNet (SL) | 59.3 |

## 4.4. Hybrid multi-task learning

HMTL in this section addressed emotion recognition from single images under both categorical and dimensional formulations. Our primary focus was the categorical model. To demonstrate the generality of HMTL, we also evaluated the circumplex model using a 3×3 puzzling auxiliary task.

For the categorical setting we considered four SSL variants:

- Puzzling (Equation 1)
- Puzzling-Rotation (Equation 2)
- In-painting-pl (Equation 3)
- In-painting-pwl

For the dimensional setting we considered three variants:

- SL regression
- SL regression-categorical (Equation 4)
- SL regression-categorical + SSL puzzling

As in prior sections, each configuration was trained three times with fixed random seeds and we reported the best validation result.

### 4.4.1. Categorical

As illustrated in Figs. 6 and 7, we attached self-supervised heads (SSHs) to the shared backbone during training of the categorical model and kept only the supervised head (SH) at evaluation. All puzzling SSHs used categorical cross-entropy as in Equation (1). We did not adopt focal loss because its values decayed more rapidly than cross-entropy, which reduced the training signal from the SSHs after a few epochs and biased optimization toward the SH. For in-painting with PL, we scaled the decoder loss by $\lambda$ so that its magnitude matched the SH loss during early training, as in Equation (3).

Due to hardware limits, HMTL experiments used random initialization with either no augmentation or weak augmentation. Although training required the network to solve both the SSL pretext (for example, predicting tile locations or reconstructing masked content) and the emotion classification, validation used only the original images without permutation or cutout, and only the SH contributed to the metric. Results are summarized in Fig. 10 and Table 5.

We note that the number of emotion categories affected accuracy. Adding Contempt, which increased the cardinality from seven to eight, reduced accuracy by roughly 3 to 4 percentage points. To reflect the harder and more general setting, we reported results for the eight-category problem throughout.

SSHs also reduced overfitting in low-data regimes. To test this, we trained HMTL using a 20 percent subset of AffectNet with the original class distribution preserved, which resulted in a heavily imbalanced subset. The SH used weighted cross-entropy. We compared SL and SL+SSL under no and weak augmentation. This experiment differed from semi-supervised learning. In semi-supervised learning, models typically pretrain on unlabeled data then fine-tune on a small labeled set, which changes both the number of images and labels. In our setting, the number of labeled images and labels remained fixed; only the training fraction varied.

We observed that removing the softmax normalization from Equation (3) destabilized training and sometimes caused collapse, especially in the low-data regime. The softmax acted as a necessary normalizer for the representation-space loss.

Table 4 reported that HMTL improved performance over SL baselines in the categorical-to-dimensional coupling and that auxiliary puzzling further improved continuous targets. To visualize where the model focused, we applied Grad-CAM [46] to representative samples (Fig. 11). With HMTL, attention concentrated more consistently on expression-relevant facial regions, which aligned with the observed gains.

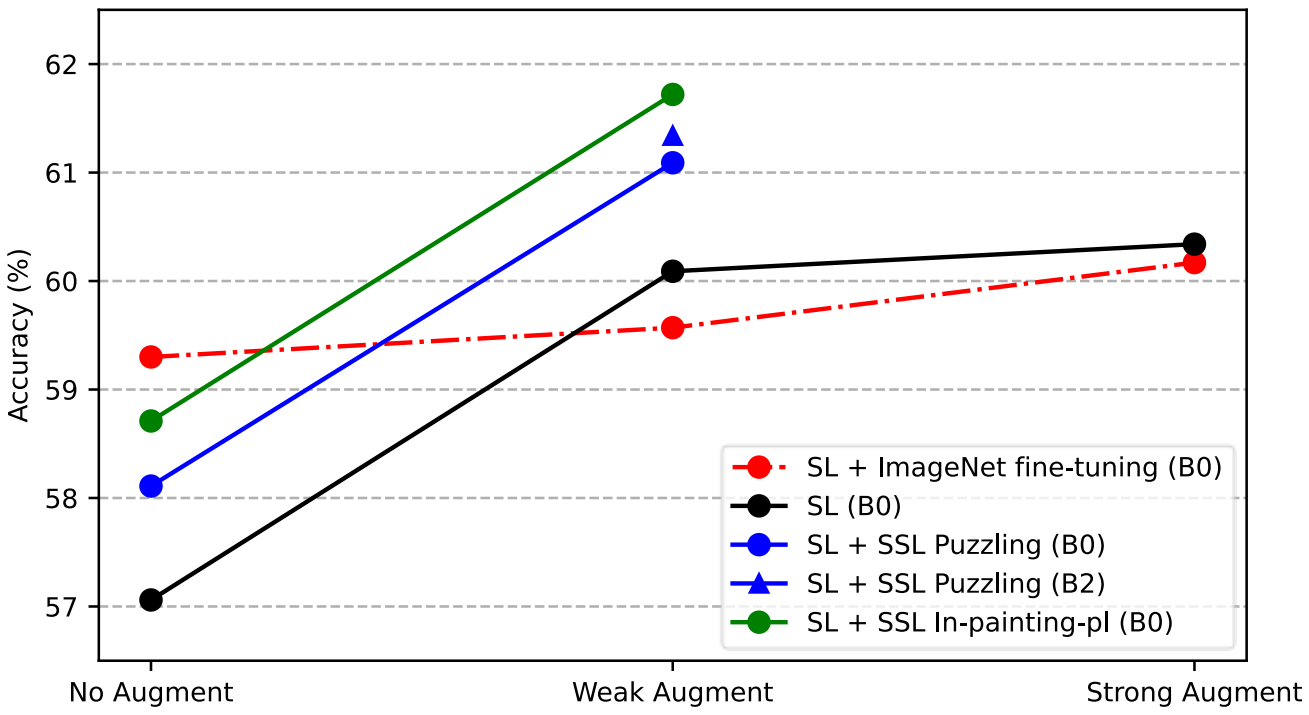

Fig. 10 Comparison of all SL and HMTL settings. The in-painting-PL configuration used cutout augmentation.

## 4.4.2. Dimensional

For the dimensional formulation, we used all AffectNet training images labeled with the eight Ekman emotions plus the None label. We set $\alpha = 1$in Equation (4) and discretized the continuous targets into 20 bins to construct categorical proxies. Table 4 reports concordance correlation coefficient (CCC) and RMSE for valence and arousal. Relative to the SL regression baseline, the regression plus categorical proxy improved both targets, and adding SSL puzzling delivered additional gains in mean CCC and RMSE.

Table 4 Results for the Russell dimensional model on the AffectNet validation set. All models were trained from random initialization. “reg-cat” denotes the categorical proxy formulation used to support regression. We report CCC and RMSE for valence and arousal and the mean across the two targets. Higher CCC and lower RMSE indicate better performance.

| Methods | Valence | | Arousal | | Mean | |
|---|---|---|---|---|---|---|
| | CCC | RMSE | CCC | RMSE | CCC | RMSE |
| ResNet50 [11] | 0.60 | 0.37 | 0.34 | 0.41 | 0.47 | 0.39 |
| SL (B0) | 0.55 | 0.39 | 0.33 | 0.41 | 0.44 | 0.43 |
| SL reg-cat (B0) | 0.57 | 0.38 | 0.42 | 0.37 | 0.49 | 0.38 |
| SL reg-cat + SSL puzzling (B0) | 0.58 | 0.38 | 0.45 | 0.36 | 0.51 | 0.37 |

### 4.5.1. Effect of puzzle sizes

We evaluated the effect of puzzle granularity on supervised head (SH) performance by comparing a no-puzzle baseline with 2×2, 3×3, and 4×4 puzzling. In the no-puzzle setting we trained only the SH. In the other settings we trained HMTL with puzzling auxiliary heads. We used the no-augmentation regime and trained all models for 20 epochs. When the puzzle size was 4×4, the emotion classifier exhibited a marked drop in accuracy. Our hypothesis was that many tiles in 4×4 configurations contained little or no expression-relevant content, which diverted learning toward incidental cues. Fig. 12 illustrated such cases where several regions carried unrelated information. To investigate further, we reweighted the losses of the self-supervised heads (SSHs) as in Equation (5).

## 4.5. Ablation Study

Feeding puzzled or cutout images to the SH without attaching SSHs could, in principle, influence performance. We therefore ablated two factors. First, we varied puzzle granularity to measure sensitivity to tile count. Second, we isolated the contribution of SSHs by comparing SH performance with and without auxiliary objectives under matched data pipelines. Table 5 presents the comparisons on the AffectNet validation set.

Table 5 Comparison between different methods on the AffectNet validation set. All of the puzzling methods are 3×3. All rotations are in eight directions. The pl refers to perceptual loss and the pwl refers to pixel-wise loss. SL+SSL refers to HMTL.

| Approach | Method | Classes | Augment level | Backbone pre-training weights | Accuracy (%) | Difference (%) |
|---|---|---|---|---|---|---|
| Supervised Learning | ResNet50 [11] | 8 | ≈Weak | - | 58.0 | -2.09 |
| | ESR-9 [47] | 8 | Unknown | - | 59.3 | -0.79 |
| | RAN [49] | 8 | Unknown | MS-Celeb-1M [48] | 59.5 | -0.59 |
| | PSR [51] | 8 | Unknown | DIV2K (STN) [50] | 60.68 | 0.59 |
| Supervised Learning | SL (B0) | 8 | No | - | 57.03 | -3.06 |
| | SL (B0) | 8 | Weak | - | 60.09 | 0.00 |
| | SL (B0) | 8 | Strong | - | 60.34 | 0.25 |
| | SL (B0) | 8 | No | AffectNet (SSL puzzling) | 57.56 | -2.53 |
| | SL (B0) | 8 | No | AffectNet (SSL rotation) | 54.26 | -5.83 |
| | SL (B0) | 8 | No | AffectNet (SSL puzzling-rotation) | 58.86 | -1.23 |
| | SL (B0) | 8 | No | ImageNet | 59.3 | -0.79 |
| | SL (B0) | 8 | Weak | ImageNet | 59.57 | -0.52 |
| | SL (B0) | 8 | Strong | ImageNet | 60.17 | 0.08 |
| | SL (B2) | 8 | Weak | ImageNet | 60.35 | 0.26 |
| Hybrid Multi-Task Learning | SL + SSL puzzling-rotation (B0) | 8 | No | - | 55.21 | -4.88 |
| | SL + SSL in-panting-pwl (B0) | 8 | No + Cutout | - | 56.78 | -3.31 |
| | SL + SSL in-panting-pl (B0) | 8 | No + Cutout | - | 58.76 | -1.33 |
| | SL + SSL in-panting-pl (B0) | 8 | Weak + Cutout | - | **61.72** | 1.63 |
| | SL + SSL puzzling (B0) | 8 | No | - | 58.11 | -1.98 |
| | SL + SSL puzzling (B0) | 8 | Weak | - | **61.09** | 1.00 |
| | SL + SSL puzzling (B2) | 8 | Weak | - | **61.32** | 1.23 |
| 20% of Training Data | SL (B0) | 8 | No | - | 43.59 | - |
| | SL (B0) | 8 | Weak | - | 52.46 | - |
| | SL+ SSL puzzling (B0) | 8 | No | - | 52.11 | - |
| | SL+ SSL puzzling (B0) | 8 | Weak | - | 54.98 | - |
| | SL+ SSL in-painting-pl (B0) | 8 | Weak + Cutout | - | 55.36 | - |

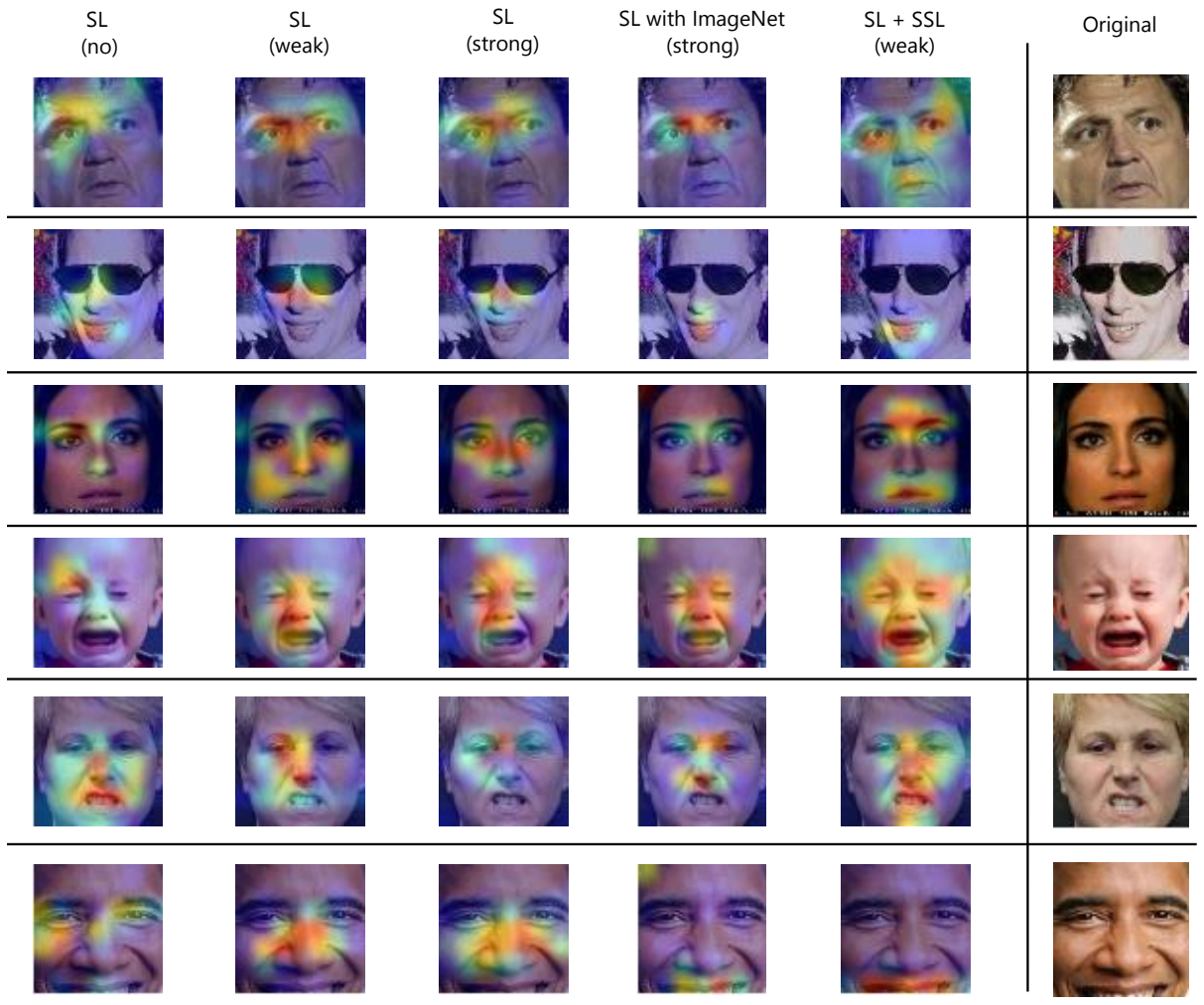

Fig.11 Grad-CAM visualizations of expression-relevant regions using the 3×3 puzzling auxiliary task. Samples were randomly drawn from the validation set. From top to bottom, the true labels were fear, happy, neutral, sad, and disgust.

$$L_{total} = \lambda_{SL} L_{SL} + \sum_{j} \lambda_j L_{SSL_j}$$

$$= -\lambda_{SL} \sum_{i} w_e y_i \log(\hat{y}_i) - \sum_{j} \lambda_j \sum_{i} y_{i,j} \log(\hat{y}_{i,j}), \qquad (5)$$

Where:

- $\lambda_{SL}$: weight for the supervised head
- $\lambda_j$: weight for each self-supervised head

During training we assigned $\lambda_j$ based on the puzzle permutations and selected values empirically. Results are shown in Fig. 13.

### 4.5.2. Effect of self-supervised heads on FER

We next removed the SSHs but continued to feed puzzled images during training. As shown in Fig. 14, the training error decreased more slowly than when SSHs were present, which made emotion recognition harder. With only the SH, the model required nearly twice as many steps to reach 50 percent validation accuracy. The gap widened as puzzle size and augmentation strength increased. Conversely, when the average SSH accuracy exceeded 80 percent, the SH loss decreased more rapidly. These observations indicated that SSHs helped the backbone learn to parse facial parts before predicting emotions, which improved downstream FER performance (Table 6).

## 4.6. Barlow Twins self-supervised learning

Recent non-contrastive SSL methods narrowed the gap to supervised learning [7,52,53]. Barlow twins [8] was a representative approach that simplified pretraining relative to methods such as BYOL and DINO. It used a Siamese architecture that encouraged two augmented views of the same image to have identical representations while reducing redundancy through a cross-correlation objective.

We pre-trained an EfficientNet-B0 backbone on the AffectNet training set using Barlow Twins under the same configuration as Section 4.2. To form two views per image, we applied the strong augmentation pipeline twice. Although Barlow Twins produced strong frozen features, it did not outperform SL baselines after fine-tuning on AffectNet labels. Table 7 compared Barlow Twins with our pretext-based SSL methods using linear evaluation and with supervised fine-tuning on AffectNet. Linear evaluation with Barlow Twins outperformed other SSL variants in the frozen setting, yet fine-tuning remained behind SL and HMTL results.

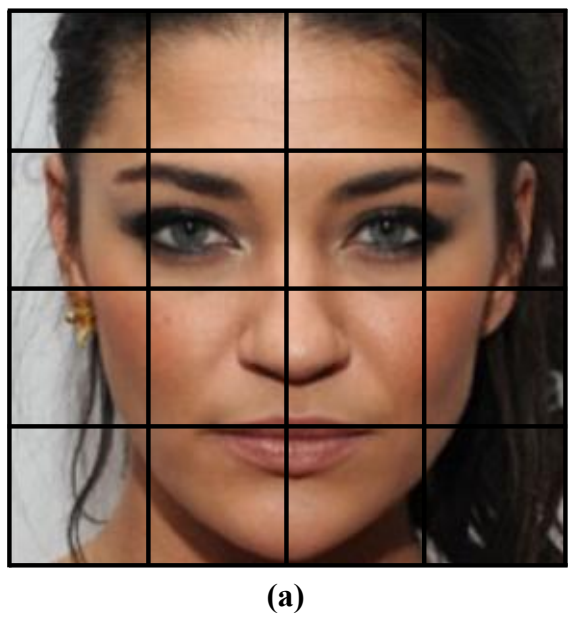

(a)

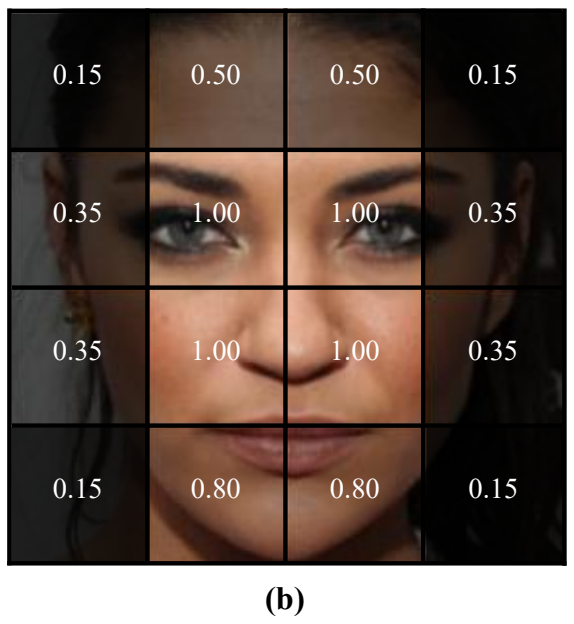

(b)

Fig. 12 Illustration of the 4×4 puzzling setting. (a) Different tiles carried different amounts of expression-relevant information, for example corner tiles often contained background. (b) We assigned region-dependent weights before puzzling to mitigate this mismatch.

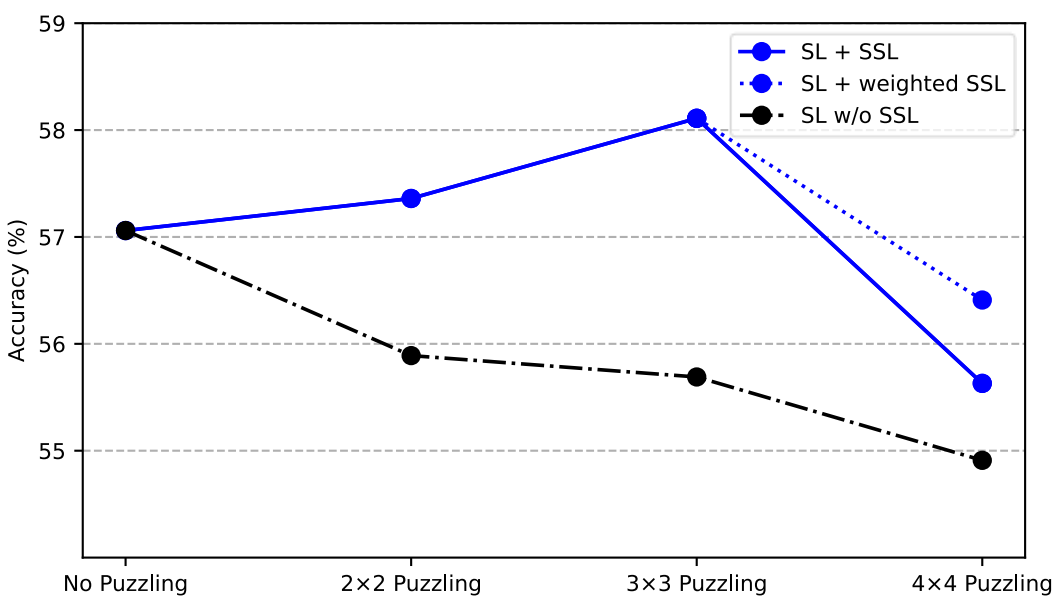

Fig. 13 Effect of puzzle size on validation accuracy. In the 4×4 setting, assigning higher weights to informative regions improved accuracy. All models were trained for 20 epochs.

Table 6 Effect of auxiliary heads. EfficientNet-B0 backbones were trained from scratch with no augmentation, except inpainting which used cutout for reconstruction. "SL + in-painting w/o SSL" was identical to "SL + cutout".

| Methods | Accuracy (%) |
|---|---|
| SL | 57.03 |
| SL + 2×2 puzzling w/o SSL | 55.89 |
| SL + 3×3 puzzling w/o SSL | 55.69 |
| SL + in-painting w/o SSL | 57.31 |
| SL + SSL 2×2 puzzling | 57.36 |
| SL + SSL 3×3 puzzling | 58.11 |
| SL + SSL in-painting | 58.71 |

Table 7 Comparison on AffectNet with EfficientNet-B0. Puzzling used a 3×3 grid. In-painting pretraining used a CNN decoder trained with a pixel-wise loss. Linear evaluation for Barlow Twins used a single linear classifier on the frozen backbone. Other SSL evaluations followed Section 4.3. Fine-tuning entries reported supervised training on AffectNet initialized from the indicated backbones.

| Backbone weights | Methods | Augment | Accuracy (%) | Macro F1 |
|---|---|---|---|---|
| Barlow Twins | Linear eval | No | 41.51 | 0.4116 |
| In-painting-pwl | Nonlinear eval | No | 34.41 | 0. 3372 |
| Puzzling | Nonlinear eval | No | 32.98 | 0.3227 |
| Random initialization | Nonlinear eval | No | 12.5 | 0.04 |
| AffectNet SSL Barlow Twins (fine-tuning) | SL | Weak | 59.11 | - |
| | SL | Strong | 59.94 | - |
| ImageNet SL (fine-tuning) | SL | Weak | 59.57 | - |
| | SL | Strong | 60.17 | - |
| Random initialization | SL | Weak | 60.09 | - |
| | SL | Strong | 60.34 | - |
| | SL + SSL puzzling | Weak | 61.09 | - |
| | SL + SSL in-panting-pl | Weak + Cutout | 61.72 | - |

## 4.7. Facial emotion recognition benchmarks

AffectNet posed a challenging benchmark due to its diversity of identities, poses, and capture conditions. A model that learned robust features on AffectNet should transfer better to other FER datasets and handle out-of-distribution examples more reliably. For cross-benchmark evaluation, we selected the backbone that was pre-trained with HMTL using 3×3 puzzling under weak augmentation from random initialization.

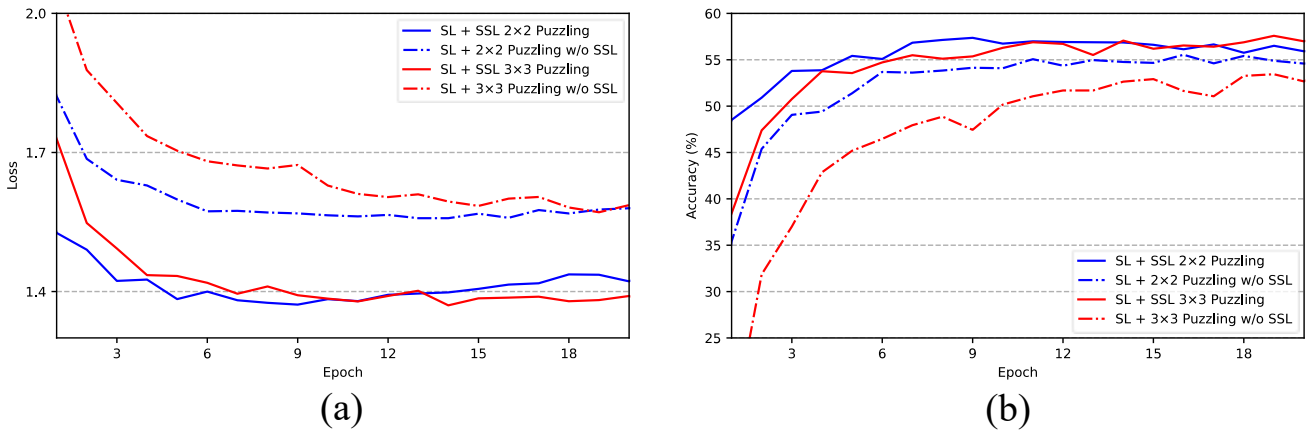

Fig. 14 Training curves under 2×2 and 3×3 puzzling. (a) loss, (b) accuracy on the AffectNet validation set. Training without SSHs progressed nearly twice as slowly to reach comparable validation accuracy.

### 4.7.1. AffWild

AffWild [54] is an in-the-wild video dataset for continuous arousal and valence. It contains more than fifteen hours of footage across 300 videos with frame-level annotations. The official training set provided 252 annotated videos. Since the challenge had concluded, we created a validation split by selecting 16 videos from the training set, as described in Appendix D. During preprocessing, we encountered an annotation issue that is detailed in Appendix D.

Following the baseline protocol [55], we cropped faces from each frame, formed sequences of length 32, and trained a two-layer bidirectional GRU with 64 units per layer. Frames were first encoded by the EfficientNet-B0 backbone to produce per-frame representations. The GRU then predicted arousal and valence per frame. We initialized the backbone with different pretraining schemes to compare transfer performance. To reduce overfitting we inserted a dropout layer with rate 0.3 after the encoder. Results are given in Table 8.

Table 8 Results on AffWild with EfficientNet-B0 backbones. HMTL pretraining used 3×3 puzzling. Both HMTL and SL pretraining were performed with weak augmentation on AffectNet. AffWild training used weak augmentation without horizontal flips.

| Backbone weights | Resolution | Valence | | Arousal | | Mean | |
|---|---|---|---|---|---|---|---|
| | | CCC | MSE | CCC | MSE | CCC | MSE |
| Random | 112×112 | 0.158 | 0.148 | 0.199 | 0.199 | 0.179 | 0.118 |
| | 224×224 | 0.268 | 0.093 | 0.245 | 0.102 | 0.256 | 0.097 |
| SL pre-training | 112×112 | 0.165 | 0.137 | 0.205 | 0.091 | 0.185 | 0.114 |
| | 224×224 | 0.281 | 0.11 | 0.292 | 0.102 | 0.286 | 0.106 |
| HMTL pre-training | 112×112 | 0.194 | 0.124 | 0.204 | 0.094 | 0.199 | 0.109 |
| | 224×224 | **0.342** | 0.09 | **0.312** | 0.103 | **0.327** | 0.097 |

### 4.7.2. AFEW-VA

AFEW-VA [56] contained video sequences annotated per frame in the circumplex space. We cropped face regions using the provided bounding boxes, then extracted fixed-length vectors with frozen backbones that were pre-trained on AffectNet: HMTL with 3×3 puzzling and a standard SL model. We trained a single-layer bidirectional LSTM with a window size of 32 to predict frame-level valence and arousal. Table 9 reports results with 10-fold cross-validation. The compared baseline was trained directly on AFEW-VA.

Table 9 Evaluation of frozen features from HMTL 3×3 puzzling and SL backbones pre-trained on AffectNet, measured on AFEW-VA with 10-fold cross-validation. Lower RMSE is better.

| Methods | Valence (RMSE) | Arousal (RMSE) |
|---|---|---|
| [56] | 0.26 | 0.22 |
| SL-eval (B0) | 0.269 | 0.252 |
| SL+ SSL puzzling-eval (B0) | 0.261 | 0.243 |

### 4.7.3. CK+

CK+ [57] comprised video sequences from 10 subjects. We cropped faces from frames, produced fixed vectors with frozen HMTL 3×3 puzzling and SL backbones pre-trained on AffectNet, and trained a single-layer bidirectional LSTM to classify sequences into seven emotions. We used 10-fold cross-validation with leave-one-subject-out. Table 10 summarizes the results and includes methods trained directly on CK+.

Table 10 Evaluation on CK+ with 10-fold cross-validation (one subject per fold). Higher accuracy is better.

| Methods | Accuracy (%) |
|---|---|
| [29] | 98 |
| [58] | 98.06 |
| SL-eval (B0) | 97.87 |
| SL+ SSL puzzling (B0)-eval | 98.23 |

### 4.7.4. JAFFE

For JAFFE, we extracted a single vector per image with the frozen HMTL 3×3 puzzling and SL backbones pre-trained on AffectNet, then trained a linear classifier over seven emotion classes. We evaluated with 10-fold cross-validation that held out one subject per fold. Table 11 lists the results and a method trained directly on JAFFE.

Table 11 Linear evaluation on JAFFE with 10-fold cross-validation. The baseline method was trained directly on the dataset.

| Methods | Accuracy (%) |
|---|---|
| [29] | 92.8 |
| SL-eval (B0) | 77.6 |
| SL+ SSL puzzling-eval (B0) | 79.88 |

## 4.8. Hybrid multi-task learning on other facial tasks

We further assessed HMTL on two facial analysis tasks: head pose estimation and gender recognition.

### 4.8.1. Hybrid multi-task learning on fine-grained head pose estimation

We evaluated head pose estimation using HopeNet [44] as the baseline and trained all methods on 300W-LP [59], which provided 61,225 synthetically posed faces. Training used random zoom, downsampling, image blur, and cutout. We reported mean absolute error for yaw, pitch, and roll on AFLW2000 [59] and followed the standard practice of removing samples with absolute angles greater than 99 degrees. To isolate the contribution of auxiliary learning, we also removed SSHs while keeping the same puzzling configuration. Figure 15 illustrates trends and Table 12 reports the errors.

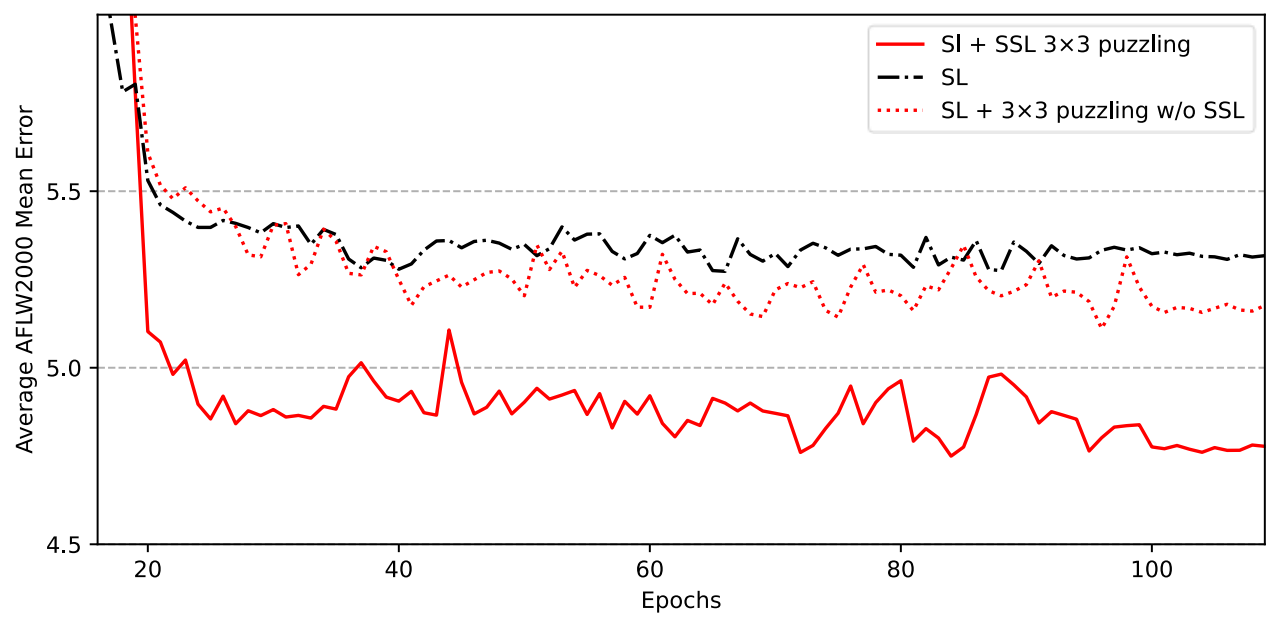

Fig. 15 Effect of auxiliary SSHs on mean angular error for head pose estimation. Curves show averages smoothed with a Gaussian filter. Raw values appear in Appendix E.

Table 12 Mean angular error on AFLW2000 after training on 300W-LP. All methods used a ResNet50 backbone for comparability with HopeNet. An asterisk marks reproduced HopeNet results.

| Method | Yaw | Pitch | Roll | Average |
|---|---|---|---|---|
| HopeNet [44] | 6.47 | 6.559 | 5.436 | 6.155 |
| SL* | 6.221 | 5.569 | 3.984 | 5.258 |
| SL + 3×3 puzzling w/o SSHs* | 4.589 | 6.223 | 4.465 | 5.092 |
| SL + SSL 3×3 puzzling | 3.898 | 5.962 | 4.479 | **4.78** |

### 4.8.2. Hybrid multi-task learning on gender recognition

We trained gender classification models on FairFace [60] a race-balanced dataset with 108,501 images spanning seven racial groups and including labels for gender, race, and age. We applied our hybrid puzzling approach at the no-augmentation level and compared to matched SL baselines. To isolate the effect of auxiliary learning, we also evaluated variants without SSHs under the same preprocessing. Table 13 reports validation accuracy on FairFace.

Table 13 Gender classification on FairFace. All models used EfficientNet-B0. Parentheses show standard deviation over three runs. The entry "SL + in-painting w/o SSL" was identical to "SL + cutout".

| Method | Accuracy (%) |
|---|---|
| SL (35 epochs) | 91.51 (±0.02) |
| SL + in-painting w/o SSL (40 epoch) | 91.59 (±0.02) |
| SL + SSL in-painting-pl | 92.12 (±0.01) |
| SL + 2×2 puzzling w/o SSL (35 epochs) | 91.33 (±0.03) |
| SL + SSL 2×2 puzzling (25 epochs) | 91.98 (±0.01) |
| SL + 3×3 puzzling w/o SSL (45 epochs) | 91.58 (±0.04) |
| SL + SSL 3×3 puzzling (35 epochs) | **92.41** (±0.01) |

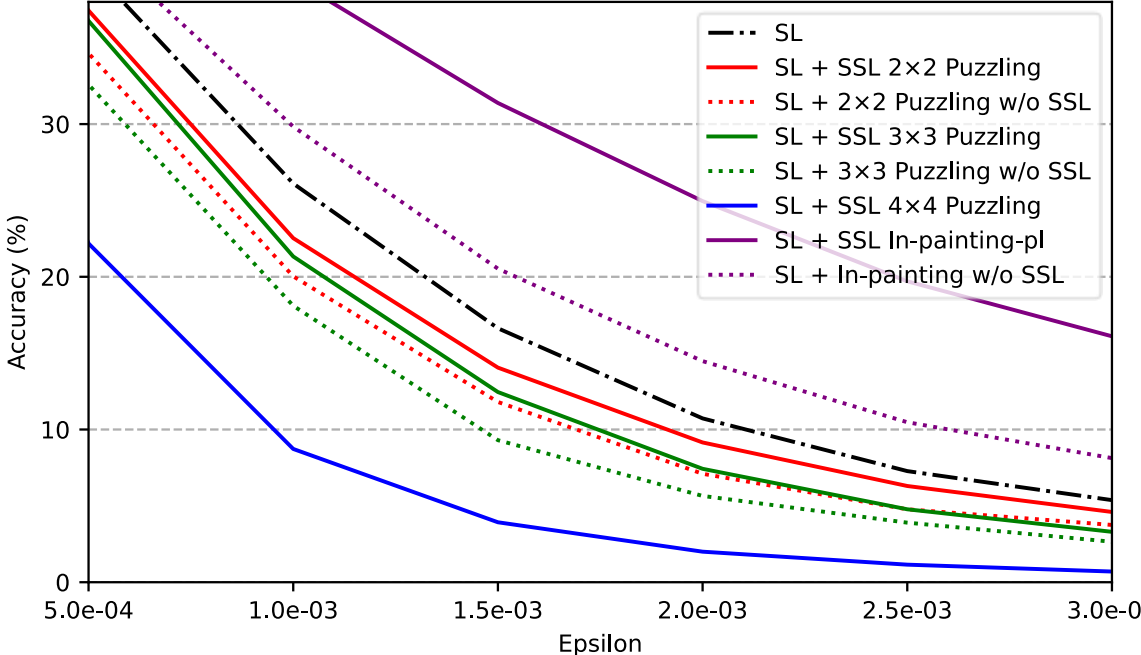

Fig. 16 Robustness on AffectNet under FGSM with varying epsilon. Models were selected by lowest validation loss. All models were trained with no augmentation. The entry "SL + in-painting w/o SSL" was identical to "SL + cutout".

## 4.6. Adversarial robustness

An adversarial attack perturbs an input image with small, human-imperceptible changes that cause a classifier to err while the original image would be correctly classified. We adopted the Fast Gradient Sign Method (FGSM) [61] (Equation 6).

$$X^{adv} = X + \varepsilon \cdot sign\left(\nabla_X J(X, y_{true})\right), \tag{6}$$

Where:

- $X^{adv}$: adversarial image
- $X$: original image
- $\varepsilon$: scale of the perturbations, by multiplying them a small float value
- $J(X, y_{true})$: the mathematical representation of the loss of the model, where X is the input to the model and y is the true label of the image

We generated FGSM adversarial examples for a range of $\varepsilon$values and evaluated models on the AffectNet validation set (Fig. 16). All backbones used EfficientNet-B0, were trained with the no-augmentation regime and dropout, and were optimized with AdaBelief. The SL+SSL 3×3 puzzling model achieved higher clean accuracy than the SL baseline, yet its accuracy degraded more rapidly under FGSM as $\varepsilon$increased. This effect became stronger with larger puzzle sizes. In contrast, adding the in-painting auxiliary head substantially improved robustness, producing markedly smaller accuracy drops across the same perturbation levels.

# 5. Discussion

In this study, we investigated three topics:

- The effect of ImageNet transfer learning versus random initialization for FER under different augmentation levels.
- The effectiveness of puzzling, rotation, and in-painting self-supervised pretraining features for FER.
- The impact of adding auxiliary self-supervised tasks during training for fine-grained facial representation learning.

## 5.1. ImageNet transfer learning vs training from scratch

With the advent of deep learning, transfer learning became widely adopted and delivered strong performance on many datasets [62,63]. In addition to random initialization, we evaluated ImageNet initialization and varied augmentation strength. The results were consistent with two observations. First, when data were limited and augmentation was weak, fine-tuning ImageNet weights improved accuracy over random initialization. Second, as augmentation became stronger, training from scratch matched or exceeded performance from ImageNet fine-tuning. A plausible explanation is that strong augmentations supplied the invariances that pretraining would otherwise provide and allowed the model to adapt its features more flexibly to FER.

## 5.2. Self-Supervised pre-training

Recent SSL approaches have reduced the gap to SL on generic benchmarks. We evaluated two SSL families on AffectNet: pretext tasks and a non-contrastive method. Under frozen-backbone evaluation, puzzling and in-painting with a pixel-wise loss produced better FER representations than rotation or the combined puzzling-rotation task. We then fine-tuned the SSL-initialized backbones. Puzzling and puzzling-rotation pretraining improved over training from scratch, while the Barlow Twins backbone performed well under linear evaluation but did not surpass SL after fine-tuning. These outcomes suggest that the benefits of SSL depend on alignment between the pretext objective and fine-grained facial cues.

## 5.3. Hybrid multi-task learning

Our HMTL hypothesis stated that adding suitable SSL auxiliary objectives during training would improve the supervised head (SH). The hypothesis held on FER and also transferred to head pose estimation and gender recognition. Auxiliary puzzling and in-painting with a perceptual loss increased accuracy and reduced error, yielding state-of-the-art results on AffectNet.

The important points in this section are as follows:

- Puzzle granularity mattered. Increasing the grid beyond 3×3 did not help and often hurt, since many 4×4 tiles contained little expression-relevant content and diverted learning.

- Overfitting was reduced across augmentation regimes. Two factors likely contributed. Multi-task optimization regularized the backbone, and puzzling injected strong perturbations. For a 3×3 grid, each image admitted 9! permutations, which expanded the effective training signal.

- Removing SSHs while still feeding puzzled inputs slowed convergence and slightly reduced accuracy relative to classical SL. The number of steps to reach the best result increased with puzzle size and with stronger augmentation. On average, switching from SL to HMTL increased the steps to the best validation result by about 30 percent, while the 3×3 setting still delivered the best final error.

- In the low-data regime with a 20 percent labeled subset that mirrored the original class imbalance, HMTL produced larger gains, especially with minimal augmentation.

## 5.4. Limitations and future works

We expect auxiliary self-supervised tasks to help beyond FER, yet our evaluation covered only three fine-grained facial benchmarks. We also faced practical issues: choosing suitable SSL objectives, setting task weights in multi-task optimization, and keeping HMTL training stable:

1. **How to select the best architecture when using self-supervised auxiliary tasks?**
   Selecting SSL objectives alongside SL requires careful screening, since different tasks interact with the supervised head (SH) in different ways. Architecture choice should be part of this screening: backbone capacity, where to attach SSL heads, and how much feature sharing to allow can change outcomes [16]. A practical procedure is: start with a small search over backbones and attachment points, run short proxy runs, compare linear probes and few-shot fine-tuning, and examine representation similarity across layers. Lightweight ablations on head capacity and normalization often reveal whether an auxiliary task cooperates or competes with the SH. Finally, treat augmentation policy and input resolution as architectural choices, because we observed strong interactions with initialization and augmentation strength.

2. **How to assign weights for different tasks in an MTL setting?**
Relative loss scales affected stability and final accuracy. In head pose estimation, the decoder loss was much smaller than the HopeNet-style supervised loss. Increasing its coefficient improved the signal early, yet too large a value destabilized training, especially in low-data regimes for in-painting with perceptual loss. Simple schedules helped in practice: warm up the SSL weight, cap it in early epochs, and reduce it once the supervised loss stabilizes. Normalizing features in the decoder loss and keeping the softmax normalization avoided collapse. More principled weighting, such as uncertainty weighting [64] or dynamic weighting [65], is a natural next step.

3. **How can SSL be most helpful for downstream tasks?**
We observed a mismatch between frozen-feature quality and fine-tuning gains. In-painting with pixel-wise loss produced strong features under nonlinear evaluation, yet fine-tuning from those weights did not surpass SL. In contrast, puzzling and puzzling plus rotation helped when fine-tuned, and puzzling or in-painting with perceptual loss helped when used as auxiliary tasks. The choice should therefore depend on how SSL is used. If the goal is frozen-feature transfer, evaluate with linear or shallow nonlinear probes. If the goal is fine-tuning or auxiliary training, judge by validation accuracy after a short fine-tuning schedule.

4. **Which SSL task should be used alongside SL?**
On AffectNet, puzzling and in-painting with a perceptual loss reduced error. Pixel-wise in-painting, rotation, and rotation plus puzzling did not consistently help and sometimes degraded performance. A data-driven selection is advisable: run brief HMTL trials, keep auxiliaries that improve the supervised validation metric, and drop those that do not. Given recent SSL diversity, extending this screening to a small pool of tasks is a practical path to demonstrate HMTL's effectiveness without changing the rest of the pipeline.

# 6. Conclusion

We conducted a comprehensive study of fine-grained FER that examined initialization, augmentation, and hybrid training with auxiliary self-supervision. First, we quantified the effect of strong augmentations and showed that, on AffectNet, training from random initialization matched or surpassed ImageNet fine-tuning once augmentation was sufficiently strong. This finding clarified when transfer learning helped and when it could be replaced by an appropriate augmentation regime. We then introduced Hybrid Multi-Task Learning, which attached self-supervised heads during training and removed them at inference. We instantiated HMTL with two auxiliary objectives, puzzling and inpainting with a perceptual loss, and demonstrated consistent gains for both categorical and dimensional FER. Across augmentation levels and in a low-data regime, HMTL improved accuracy and reduced error. With two HMTL variants, we achieved state-of-the-art performance on AffectNet in the eight-emotion setting without using additional pretraining data.

Ablations provided practical guidance. Puzzle granularity mattered, 3×3 delivered the best trade-off while 4×4 often diluted expression cues. Removing auxiliary heads while keeping puzzled inputs slowed convergence and slightly reduced peak accuracy, which suggested that the auxiliary objectives shaped useful intermediate features rather than acting only as data augmentation. Robustness experiments indicated complementary behavior, puzzling improved clean accuracy but degraded more under FGSM as the grid size grew, while inpainting with a perceptual loss substantially improved adversarial robustness.

We also assessed transfer beyond FER. HMTL improved fine-grained head pose estimation and gender recognition, and frozen features transferred competitively to AFEW-VA, CK+, and JAFFE. These results supported the view that auxiliary self-supervision encouraged part-aware, expression-relevant representations that generalize across tasks and datasets.

Finally, our analysis of SSL usage highlighted an important nuance. SSL pretraining alone did not guarantee fine-tuning gains, for example Barlow Twins excelled under linear evaluation but did not surpass supervised baselines after full fine-tuning on AffectNet. The greatest benefits arose when SSL acted as an auxiliary objective during supervised training and was aligned with the downstream target.

Overall, HMTL offered a simple and effective recipe for improving fine-grained facial analysis: keep supervision unchanged, add aligned self-supervised heads during training, and remove them at test time. Future work will automate task selection and weighting, extend the approach to temporal settings and larger backbones, and test broader visual domains where subtle local structure drives recognition.

# Acknowledgments

We gratefully acknowledge Hadi Pourmirzaei for designing all pipelines pictures. Moreover, thanks to Cyrus Kazemirad for partially supporting us for hardware resources and also for proofreading and helpful discussions.

# References

1. Ko BC. A brief review of facial emotion recognition based on visual information. Sensors (Switzerland). 2018 Feb,:18(2):401.
2. Yu Z, Zhang C. Image based static facial expression recognition with multiple deep network learning. In: ICMI 2015 - Proceedings of the 2015 ACM International Conference on Multimodal Interaction. 2015.
3. Russell JA. A circumplex model of affect. J Pers Soc Psychol. 1980 Dec;3.
4. Ekman P, Friesen W V. Constants across cultures in the face and emotion. J Pers Soc Psychol. 1971;17(2):124.
5. Remington NA, Fabrigar LR, Visser PS. Reexamining the circumplex model of affect. J Pers Soc Psychol. 2000;79(2):286.
6. Chen T, Kornblith S, Swersky K, Norouzi M, Hinton G. Big Self-Supervised Models are Strong Semi-Supervised Learners. Advances in neural information processing systems, 33, pp.22243-22255. 2020.
7. Grill JB, Strub F, Altché F, Tallec C, Richemond PH, Buchatskaya E, et al. Bootstrap your own latent a new approach to self-supervised learning. Adv Neural Inf Process Syst. 2020;33:21271–84.
8. Zbontar J, Jing L, Misra I, LeCun Y, Deny S. Barlow Twins: Self-Supervised Learning via Redundancy Reduction. arXiv Prepr arXiv210303230. 2021;
9. Caron M, Touvron H, Misra I, Jégou H, Mairal J, Bojanowski P, et al. Emerging properties in self-supervised vision transformers. In: Proceedings of the IEEE/CVF International Conference on Computer Vision. 2021. p. 9650–60.
10. Cole E, Yang X, Wilber K, Mac Aodha O, Belongie S. When does contrastive visual representation learning work? arXiv Prepr arXiv210505837. 2021;
11. Mollahosseini A, Hasani B, Mahoor MH. Affectnet: A database for facial expression, valence, and arousal computing in the wild. IEEE Trans Affect Comput. 2017;10(1):18–31.
12. Gidaris S, Singh P, Komodakis N. Unsupervised representation learning by predicting image rotations. arXiv preprint arXiv:1803.07728. 2018.
13. Kim D, Cho D, Yoo D, Kweon IS. Learning image representations by completing damaged jigsaw puzzles. In: 2018 IEEE Winter Conference on Applications of Computer Vision (WACV). IEEE; 2018. p. 793–802.
14. Asano YM, Rupprecht C, Vedaldi A. A critical analysis of self-supervision, or what we can learn from a single image. arXiv Prepr arXiv190413132. 2019;
15. Standley T, Zamir A, Chen D, Guibas L, Malik J, Savarese S. Which tasks should be learned together in multi-task learning? In: International Conference on Machine Learning. PMLR; 2020. p. 9120–32.
16. Crawshaw M. Multi-task learning with deep neural networks: A survey. arXiv preprint arXiv:2009.09796. 2020.
17. Walecki R, Pavlovic V, Schuller B, Pantic M. Deep structured learning for facial action unit intensity estimation. In: Proceedings of the IEEE Conference on Computer Vision and Pattern Recognition. 2017. p. 3405–14.
18. Georgescu M-I, Ionescu RT, Popescu M. Local learning with deep and handcrafted features for facial expression recognition. IEEE Access. 2019;7:64827–36.
19. Acharya D, Huang Z, Pani Paudel D, Van Gool L. Covariance pooling for facial expression recognition. In: Proceedings of the IEEE Conference on Computer Vision and Pattern Recognition

Workshops. 2018. p. 367–74.
20. Agrawal A, Mittal N. Using CNN for facial expression recognition: a study of the effects of kernel size and number of filters on accuracy. Vis Comput. 2020;36(2):405–12.
21. Zhou F, Kong S, Fowlkes CC, Chen T, Lei B. Fine-grained facial expression analysis using dimensional emotion model. Neurocomputing. 2020;392:38–49.
22. Breuer R, Kimmel R. A deep learning perspective on the origin of facial expressions. arXiv Prepr arXiv170501842. 2017;
23. Hasani B, Mahoor MH. Facial expression recognition using enhanced deep 3D convolutional neural networks. In: Proceedings of the IEEE conference on computer vision and pattern recognition workshops. 2017. p. 30–40.
24. Jung H, Lee S, Yim J, Park S, Kim J. Joint fine-tuning in deep neural networks for facial expression recognition. In: Proceedings of the IEEE international conference on computer vision. 2015. p. 2983–91.
25. Kollias D, Zafeiriou S. Exploiting multi-cnn features in cnn-rnn based dimensional emotion recognition on the omg in-the-wild dataset. IEEE Trans Affect Comput. 2020;12(3):595–606.
26. Ebrahimi Kahou S, Michalski V, Konda K, Memisevic R, Pal C. Recurrent neural networks for emotion recognition in video. In: Proceedings of the 2015 ACM on international conference on multimodal interaction. 2015. p. 467–74.
27. Sun W, Zhao H, Jin Z. A visual attention based ROI detection method for facial expression recognition. Neurocomputing. 2018;296:12–22.
28. Wang C, Hu R, Hu M, Liu J, Ren T, He S, et al. Lossless attention in convolutional networks for facial expression recognition in the wild. arXiv Prepr arXiv200111869. 2020;
29. Minaee S, Minaei M, Abdolrashidi A. Deep-emotion: Facial expression recognition using attentional convolutional network. Sensors. 2021;21(9):3046.
30. Meng D, Peng X, Wang K, Qiao Y. Frame attention networks for facial expression recognition in videos. In: 2019 IEEE international conference on image processing (ICIP). IEEE; 2019. p. 3866–70.
31. Li Y, Zeng J, Shan S, Chen X. Occlusion aware facial expression recognition using CNN with attention mechanism. IEEE Trans Image Process. 2018;28(5):2439–50.
32. Wu C, Chai L, Yang J, Sheng Y. Facial Expression Recognition using Convolutional Neural Network on Graphs. In: 2019 Chinese Control Conference (CCC). IEEE; 2019. p. 7572–6.
33. Cai J, Meng Z, Khan AS, O'Reilly J, Li Z, Han S, et al. Identity-free facial expression recognition using conditional generative adversarial network. In: 2021 IEEE International Conference on Image Processing (ICIP). IEEE; 2021. p. 1344–8.
34. Jing L, Tian Y. Self-supervised visual feature learning with deep neural networks: A survey. IEEE Trans Pattern Anal Mach Intell. 2020;43(11):4037–58.
35. Falcon W, Cho K. A framework for contrastive self-supervised learning and designing a new approach. arXiv Prepr arXiv200900104. 2020;
36. Pathak D, Krahenbuhl P, Donahue J, Darrell T, Efros AA. Context Encoders: Feature Learning by Inpainting. In: Proceedings of the IEEE Computer Society Conference on Computer Vision and Pattern Recognition. 2016.
37. Noroozi M, Favaro P. Unsupervised learning of visual representations by solving jigsaw puzzles. In: European conference on computer vision. Springer; 2016. p. 69–84.
38. Wei C, Xie L, Ren X, Xia Y, Su C, Liu J, et al. Iterative reorganization with weak spatial constraints: Solving arbitrary jigsaw puzzles for unsupervised representation learning. In: Proceedings of the IEEE/CVF Conference on Computer Vision and Pattern Recognition. 2019. p. 1910–9.
39. Kaur H, Pannu HS, Malhi AK. A systematic review on imbalanced data challenges in machine learning: Applications and solutions. ACM Comput Surv. 2019;52(4):1–36.
40. Lin TY, Goyal P, Girshick R, He K, Dollar P. Focal Loss for Dense Object Detection. IEEE Trans Pattern Anal Mach Intell. 2020;
41. Tan M, Le Q. Efficientnet: Rethinking model scaling for convolutional neural networks. In: International conference on machine learning. PMLR; 2019. p. 6105–14.
42. Hu G, Liu L, Yuan Y, Yu Z, Hua Y, Zhang Z, et al. Deep multi-task learning to recognise subtle facial expressions of mental states. In: Proceedings of the European Conference on Computer Vision (ECCV). 2018. p. 103–19.
43. Hung SCY, Lee J-H, Wan TST, Chen C-H, Chan Y-M, Chen C-S. Increasingly packing multiple facial-informatics modules in a unified deep-learning model via lifelong learning. In: Proceedings of the 2019 on International Conference on Multimedia Retrieval. 2019. p. 339–43.
44. Ruiz N, Chong E, Rehg JM. Fine-grained head pose estimation without keypoints. In: Proceedings of the IEEE conference on computer vision and pattern recognition workshops. 2018. p. 2074–83.
45. Zhuang J, Tang T, Ding Y, Tatikonda S, Dvornek N, Papademetris X, et al. AdaBelief optimizer: Adapting stepsizes by the belief in observed gradients. Advances in neural information processing systems, 33, pp.18795-18806. 2020.
46. Selvaraju RR, Cogswell M, Das A, Vedantam R, Parikh D, Batra D. Grad-cam: Visual explanations from deep networks via gradient-based localization. In: Proceedings of the IEEE international conference on computer vision. 2017. p. 618–26.
47. Siqueira H, Magg S, Wermter S. Efficient facial feature learning with wide ensemble-based convolutional neural networks. In: Proceedings of the AAAI conference on artificial intelligence. 2020. p. 5800–9.
48. Guo Y, Zhang L, Hu Y, He X, Gao J. Ms-celeb-1m: A dataset and benchmark for large-scale face recognition. In: European conference on computer vision. Springer; 2016. p. 87–102.
49. Wang K, Peng X, Yang J, Meng D, Qiao Y. Region attention networks for pose and occlusion robust facial expression recognition. IEEE Trans Image Process. 2020;29:4057–69.
50. Agustsson E, Timofte R. Ntire 2017 challenge on single image super-resolution: Dataset and study. In: Proceedings of the IEEE Conference on Computer Vision and Pattern Recognition Workshops. 2017. p. 126–35.
51. Vo T-H, Lee G-S, Yang H-J, Kim S-H. Pyramid with super resolution for in-the-wild facial expression recognition. IEEE Access. 2020;8:131988–2001.
52. Caron M, Touvron H, Misra I, Jégou H, Mairal J, Bojanowski P, et al. Emerging Properties in Self-Supervised Vision Transformers. Proc IEEE/CVF Int Conf Comput Vis (pp 9650-9660). 2021 Apr;
53. Baevski A, Hsu W-N, Xu Q, Babu A, Gu J, Auli M. Data2vec: A general framework for self-supervised learning in speech, vision and language. arXiv Prepr arXiv220203555. 2022;
54. Kollias D, Tzirakis P, Nicolaou MA, Papaioannou A, Zhao G, Schuller B, et al. Deep affect prediction in-the-wild: Aff-wild database and challenge, deep architectures, and beyond. Int J Comput Vis. 2019;127(6):907–29.
55. Zafeiriou S, Kollias D, Nicolaou MA, Papaioannou A, Zhao G, Kotsia I. Aff-wild: valence and arousal'In-the-Wild'challenge. In: Proceedings of the IEEE conference on computer vision and pattern recognition workshops. 2017. p. 34–41.
56. Kossaifi J, Tzimiropoulos G, Todorovic S, Pantic M. AFEW-VA database for valence and arousal estimation in-the-wild. Image Vis Comput. 2017;65:23–36.
57. Lucey P, Cohn JF, Kanade T, Saragih J, Ambadar Z, Matthews I. The extended cohn-kanade dataset (ck+): A complete dataset for action unit and emotion-specified expression. In: 2010 ieee computer society conference on computer vision and pattern recognition-workshops. IEEE; 2010. p. 94–101.
58. Chen Y, Wang J, Chen S, Shi Z, Cai J. Facial motion prior networks for facial expression recognition. In: 2019 IEEE Visual Communications and Image Processing (VCIP). IEEE; 2019. p. 1–4.
59. Zhu X, Lei Z, Liu X, Shi H, Li SZ. Face alignment across large poses: A 3d solution. In: Proceedings of the IEEE conference on computer vision and pattern recognition. 2016. p. 146–55.
60. Kärkkäinen K, Joo J. Fairface: Face attribute dataset for balanced race, gender, and age. arXiv Prepr arXiv190804913. 2019;
61. Goodfellow IJ, Shlens J, Szegedy C. Explaining and harnessing adversarial examples. arXiv Prepr arXiv14126572. 2014;
62. Tan C, Sun F, Kong T, Zhang W, Yang C, Liu C. A survey on deep transfer learning. In: International conference on artificial neural networks. Springer; 2018. p. 270–9.
63. Ng H-W, Nguyen VD, Vonikakis V, Winkler S. Deep learning for emotion recognition on small datasets using transfer learning. In: Proceedings of the 2015 ACM on international conference on multimodal interaction. 2015. p. 443–9.
64. Cipolla R, Gal Y, Kendall A. Multi-task Learning Using Uncertainty to Weigh Losses for Scene Geometry and Semantics. In: Proceedings of the IEEE Computer Society Conference on Computer Vision and Pattern Recognition. 2018.
65. Ming Z, Xia J, Luqman MM, Burie J-C, Zhao K. Dynamic multi-task learning for face recognition with facial expression. arXiv Prepr

arXiv191103281. 2019;

# Appendices

## A. Augmentation details

We defined three augmentation regimes using nine transformations. Magnitudes were sampled uniformly at random within the specified ranges for each transformation. Table A1 lists the settings.

Table A1 Augmentation settings.

| Transformation | Level 1 (No augment) | Level 2 (Weak augment) | Level 3 (Strong augment) |
|---|---|---|---|
| Horizontal flip | ✓ | ✓ | ✓ |
| Central Zoom | ✕ | ✓ (0.69 to 100) | ✓ (0.69 to 100) |
| Contrast | ✕ | ✓ (0.6 to 1.4) | ✓ (0.6 to 1.4) |
| Rotation | ✕ | ✓ (-15° to 15°) | ✓ (-20° to 20°) |
| Brightness | ✕ | ✕ | ✓ (-0.05 to 0.05) |
| RGB channel swap | ✕ | ✕ | ✓ |
| Blurring | ✕ | ✕ | ✓ (1, 3, 5 filter size) |
| Gaussian noise | ✕ | ✕ | ✓ (mean=0, var=0.05) |
| Cutout | ✕ | ✕ | ✓ (60×60) |

## B. Label smoothing

To assess the effect of label smoothing, we performed an experiment on AffectNet with class balancing by down sampling to 10 percent of the training data. We used EfficientNet-B0, weak augmentation, and dropout 0.2. Table B1 reports validation accuracy. Label smoothing did not improve accuracy for emotion recognition in this setting.

Table B1 Label smoothing on AffectNet validation.

| Label smoothing | Accuracy (%) |
|---|---|
| 0.0 | 53.11 |
| 0.1 | 52.74 |
| 0.2 | 51.65 |
| 0.3 | 51.62 |

## C. Architecture details of self-supervised heads

In all HMTL configurations, the supervised emotion head was a linear classifier attached to the backbone output after global average pooling, except for the inpainting variants which used a decoder head.

**Puzzling.** Each puzzling head was a linear classifier on the global average pooling output of EfficientNet. All puzzling loss weights were set to one.

**Rotation.** The rotation head was a single linear classifier on the global average pooling output with eight classes. Its loss weight was set to one.

**Puzzling-Rotation.** Using only linear heads for both tasks caused a large drop in supervised accuracy. To prevent this, we inserted two shallow DNN towers with one hidden layer of 512 units each on top of the global average pooling output, one tower for puzzling and one for rotation. The puzzling tower then branched into the per-tile linear classifiers.

**In-painting.** The inpainting branch consisted of a five-block deconvolutional decoder with skip connections. Each block contained Conv2DTranspose, batch normalization, Conv2DTranspose, batch normalization, followed by two upsampling layers. The blocks used 256, 128, 64, 32, and 16 filters respectively. A final 1×1 convolution mapped the output to three channels.

## D. AffWild

AffWild contained 300 videos, with 252 provided as the training split and the remainder as the test split. Since the challenge had concluded, test annotations were not available, so we created a validation split by selecting the following 16 videos from the training set: 110, 179, 189, 203, 221, 249, 260, 306, 330, 332, 345, 402, 415, 433, 448, 449.

Although the dataset provided bounding boxes and landmarks, we found many of them inaccurate or noisy. In frames with multiple faces, the annotations sometimes switched identities across frames. Correcting all labels would require substantial engineering and manual work. Given the frame resolution of 640×360, processing uncropped frames would have been computationally expensive, which motivated the use of face crops.

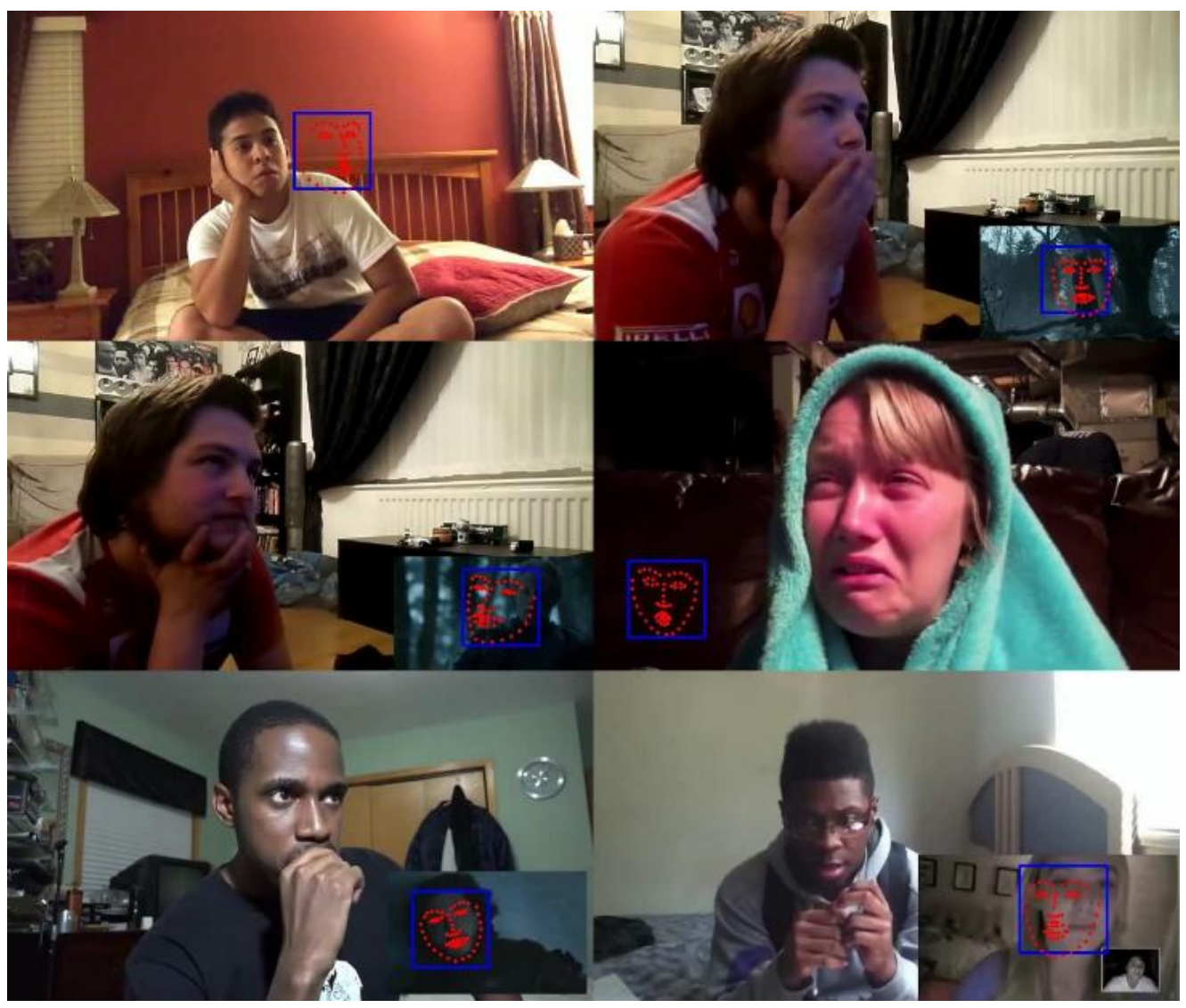

Fig. 1 AffWild original face annotations. Many frames contained incorrect bounding boxes and landmarks.

To mitigate annotation noise, we retained the original labels and re-detected faces with RetinaFace to obtain improved bounding boxes. The multiple-face issue occasionally persisted, so we selected the largest face as the target bounding box.

## E. Head pose estimation validation error rate through training epochs

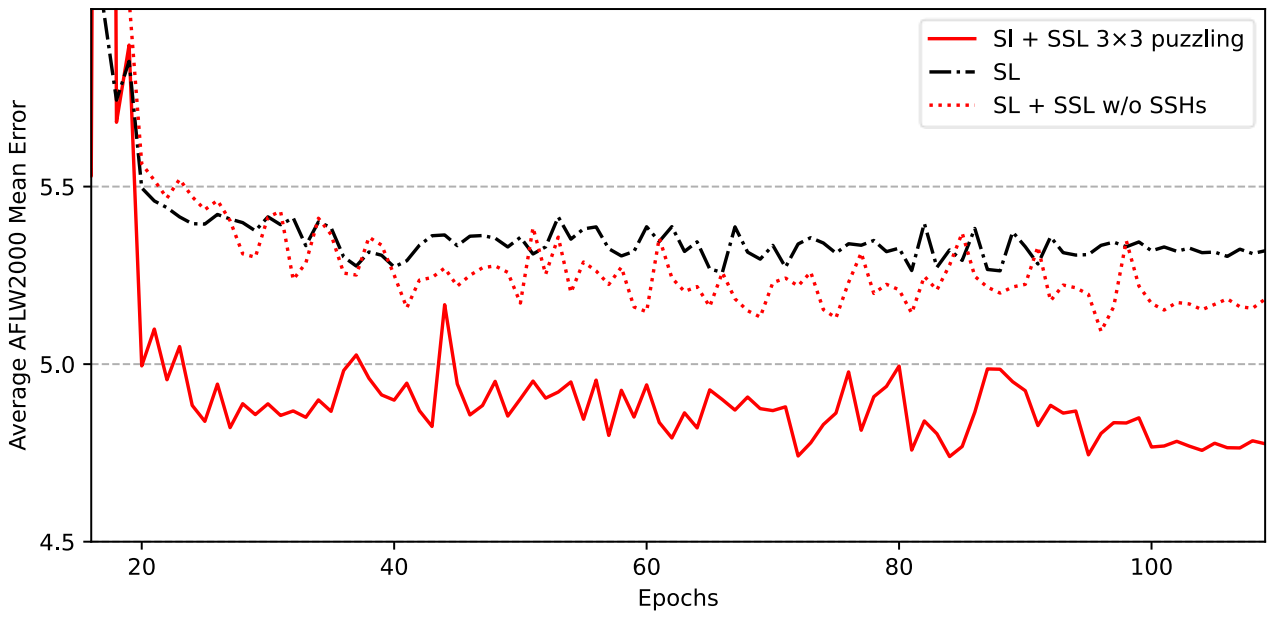

Fig. 2 Average validation error for head pose estimation across training epochs without smoothing. The main text reports smoothed curves for readability, while this figure shows the raw trajectories to facilitate comparison of SSH effects at different puzzle sizes.